\documentclass[preprint,12pt]{elsarticle}

\usepackage{amssymb}

\usepackage{amsmath,amsfonts}
\usepackage{algorithmic}
\usepackage{algorithm}
\usepackage{array}
\usepackage[caption=false,font=normalsize,labelfont=sf,textfont=sf]{subfig}
\usepackage{textcomp}
\usepackage{stfloats}
\usepackage{url}
\usepackage{verbatim}
\usepackage{graphicx}

\usepackage[linesnumbered,ruled,vlined,boxed,commentsnumbered,algo2e]{algorithm2e}

\usepackage{amsfonts,amssymb,amstext,verbatim,amsmath,color}
\usepackage{mathtools}
\usepackage{multirow}
\usepackage{xcolor}
\usepackage{float}
\usepackage{natbib}
\usepackage{adjustbox}
\usepackage[symbol]{footmisc}
\usepackage{xparse}%

\def\b#1{\mathchoice{\hbox{\boldmath $\displaystyle #1$}}
        {\hbox{\boldmath $\textstyle #1$}}
        {\hbox{\boldmath $\scriptstyle #1$}}
        {\hbox{\boldmath $\scriptscriptstyle #1$}}}

\newcommand{\preset}[1]{\ensuremath{\,\!^\bullet{#1}}}
\newcommand{\postset}[1]{\ensuremath{{#1}^\bullet}}

\newcommand{\N}{{\ensuremath{\mathcal{N}}}}    
\newcommand{\m}{\b{m}}                      

\newtheorem{theorem}{\textbf{Theorem}}[section]

\newtheorem{example}[theorem]{\textbf{Example}}

\newtheorem{definition}[theorem]{\textbf{Definition}}

\newcommand{\RomanNumeralCaps}[1]{\MakeUppercase{\romannumeral #1}}

\journal{Robotics and Autonomous Systems}

\begin{document}

\begin{frontmatter}



\title{Multi-robot Motion Planning based on  Nets-within-Nets Modeling and Simulation}



\author{Sofia Hustiu$^{a}$\footnote[1]{Corresponding author. \textit{E-mail} address: \url{sofia.hustiu@academic.tuiasi.ro}}, Joaquín Ezpeleta$^b$, Cristian Mahulea$^b$ and Marius Kloetzer$^a$ } 

\affiliation{organization={Dept. of Automatic Control and Applied Informatics, ''Gheorghe Asachi'' Technical University },
            city={Iasi},
            postcode={700050},
            country={Romania}}

            
\affiliation{organization={Aragón Institute of Engineering Research (I3A), University of Zaragoza},
            city={Zaragoza},
            postcode={50009},
            country={Spain}}

\begin{abstract}
This paper focuses on designing motion plans for a heterogeneous team of robots that must cooperate to fulfill a global mission. Robots move in an environment that contains some regions of interest, while the specification for the entire team can include avoidance, visits, or sequencing of these regions of interest. The mission is expressed in terms of a Petri net corresponding to an automaton, while each robot is also modeled by a state machine Petri net. The current work brings about the following contributions with respect to existing solutions for related problems. First, we propose a novel model, denoted \emph{High-Level robot team Petri Net} (HLrtPN) system, to incorporate the specification and robot models into the Nets-within-Nets paradigm. A guard function, named \emph{Global Enabling Function}, is designed to synchronize the firing of transitions so that robot motions do not violate the specification. Then, the solution is found by simulating the HLrtPN system in a specific software tool that accommodates Nets-within-Nets. Illustrative examples based on Linear Temporal Logic missions support the computational feasibility of the proposed framework. \end{abstract}



\begin{highlights}
\item Defining top-down formal definition of a novel framework based on the Nets-within-Nets paradigm.
\item Assessing the framework considering planning solutions for heterogeneous and homogeneous multi-agent systems.
\item The planning strategy ensures a global high-level mission.
\item Introducing step-by-step implementation guideline of the framework in Renew.
\end{highlights}

\begin{keyword}

Discrete event systems; Motion planning; Multi-robot system




\end{keyword}

\end{frontmatter}



\section{Introduction}

Planning motion in the robotics field requires collision-free navigation in an environment (known, partially, or fully unknown), ensuring imposed performances and/or fulfilling a desired mission. Suppose that simple scenarios express the mission as reaching a desired position \cite{LAV06}. In that case, complex scenarios include multi-robot systems with the mission expressed under spatial and/or temporal constraints, through various formal languages. Several examples include classes of Temporal Logic formulas \cite{lindemann2017robust,mehdipour2020specifying}, Boolean expressions \cite{yu2022security}, $\mu$-calculus specifications \cite{plaku2016motion}. An example of a complex mission can be given as \textit{the pick-up region A should be visited simultaneously by two robots (also addressed as agents). Subsequently, region B should be visited and region C should be reached immediately after B. Throughout the trajectory, region D is always avoided.} Obviously, robots should work cooperatively, ensuring the sequencing of events based on the tasks that will be assigned to them. Another example \cite{tabuada2006linear} uses Temporal Logic to specify objectives to control linear systems.


A modeling solution for the stated problem is based on formal methods including Discrete Event Systems (DES) \cite{cassandras2008introduction}, due to their well-documented properties to analyze, design, and control the behavior of a system. These representations are used in the literature for the dynamic of the robotic team in the workspace, e.g., transition systems (TS), alongside representations of the team's assignment, e.g., B{\"u}chi automaton (BA) in the case of Linear Temporal Logic (LTL) \cite{Belta-RAM07,  lacerda2019petri}. When combining the set of TS modeling the movement of each robot with the automaton of the mission, a synchronous product is computed in the search for a motion plan. The downside is represented by the exponential growth of states of the resulting product when the number of agents on the team increases. In the case of homogeneous robotic teams, Petri net (PN) models proved to be efficient in representing the abstraction of the team, by maintaining the same topology representation regardless of the team size, since a token represents one robot \cite{mahulea2020path}.  

One challenge for the stated problem appears when the multi-robot system is composed of heterogeneous robots, as needed by daily applications, e.g., in the agriculture domain \cite{ju2021hybrid}. We propose a solution based on a hierarchical approach of Petri net models, known as the Nets-within-Nets (NwN) paradigm \cite{jensen2012high}. The particularity of this family of high-level nets is characterized by the fact that each token can transfer information such as states of another process. In this sense, a token is visualized as a Petri net denoted as \textit{Object net}. Moreover, the relation between these nets is captured in \textit{System net}, which contains a global view of the entire system \cite{valk2003object}. 

This paper aims to introduce a framework called the \emph{High-Level robot team Petri Net} system HLrtPN for trajectory planning (expressed as robotic trajectories) a heterogeneous multi-robot system ensuring a global mission. The behavior of the robots, alongside a model of the mission of the team, is represented by object nets, their interaction being coordinated by the system net through a \emph{Global Enabling Function (GEF)} that has to be properly designed. A motion planning solution is obtained through simulations of the HLrtPN in a specific NwN software until the mission of the team is satisfied. The implementation of the proposed framework is accessible on our website \cite{NwNsite}, accompanied by representative examples with impact in real life, such as robot coordination within a hospital. Based on our knowledge, this is the first work to propose a step-by-step framework and motion plan solution based on the NwN paradigm.

The structure of the paper includes the following: a short literature review of the studies related to formal methods for a team of robots, focusing on Petri net models and the Nets-within-Nets paradigm, with an emphasis on the contributions of the current work (Section \ref{sec:RW}), an introduction of the problem formulation and the used notation (Section \ref{sec:pbstat}), a characterization of the proposed formalism under the Nets-within-Nets paradigm (Sections \ref{sec:objectnet}, \ref{sec:systemnet}), and a presentation of the setup containing two case studies evaluating the performances of the proposed method in comparison with other DES approaches from literature (Section \ref{sec:results}). The last section concludes the paper and provides a future perspective based on several improvements that can be further investigated for the foundation framework proposed in this paper.

\section{Related work}\label{sec:RW}

Multiple planning solutions have been proposed in the state-of-the-art for a multi-robot system, as stated in \cite{madridano2021trajectory, antonyshyn2023multiple}, including classical approaches, e.g., sampling-based algorithms; bioinspired algorithms, e.g., Genetic Algorithms; mathematical programmings, e.g., optimization problems. These methods are versatile, computing paths for Unmanned Ground Vehicles (UGV) and Unmanned Aerial Vehicles (UAV) in 2D and 3D spaces. Typically, a high-level mission assigned to a team involves reaching a set of final positions within specified space and time constraints. Thus, the aforementioned planning methods ensure a collision-free motion for the robots accomplishing the mission. In this paper, we extend the capabilities of robotic systems to fulfill missions that require the completion of tasks in a specific order or with certain dependencies. One approach to represent these complex missions is by symbolic representation, e.g., LTL formulae.

One advantage of LTL specifications is enhanced by its representation as an automaton that is easily composed with a set of TS allocated to each robot. This model facilitates the solution (robots' paths) through the use of graph-based search algorithms, in known \cite{ding2011automatic} or unknown \cite{kloetzer2015ltl} spaces. Other benefits of the tuple LTL and TS are encountered in various works, for optimally allocating tasks based on pruning the B\"uchi automaton of the LTL \cite{luo2022temporal}, for supervising the parallel composition of the entire model for the robotic system through a feedback controller \cite{ju2019modeling}, being further extended to a hybrid system (UAV and UGV) ensuring a set of behavioral policies in the agriculture field \cite{ju2021hybrid}.

TS-based approaches include one downside expressed by an aggressive growth in the size of the model with the increase of the number of robots in the team. Thus, the solution space becomes too large to be computationally tractable. This limitation of the state explosion problem can be solved through another DES representation, i.e., a PN model for a partitioned environment, maintaining its size w.r.t. the number of robots in the team. By representing each robot as a token, this approach is generally suitable for homogeneous teams \cite{mahulea2020path}, while the robotic paths are computed by optimization problems \cite{hustiu2024multi}.

Different works focused on investigating the relation between multiple PN in order to automatically translate the specification into a controller code, which can be further used for planning solutions. Specifically, in \cite{figat2022parameterised}, a representation is proposed, denoted Robotic System Hierarchical Petri Net Meta-model, consisting of several layers of PN, specifically: robotic system layer, agent layer, subsystem layer, behavior layer, and action layer.

One barrier is represented by the coordination of heterogeneous robots while preserving the advantages of PN models. One might suggest the use of different classes of PN, e.g., colored Petri nets \cite{allison2022modeling}. Other works focus on providing a framework encoding robot capabilities, task specification, and cooperative task execution into one representation, considering a stochastic environment \cite{schillinger2021adaptive}.

 Supervisory control approach has also been considered in the robotic field, as detailed in \cite{cassandras2008introduction} and extended in \cite{tabuada2006linear} for hybrid systems. The main idea here is to avoid deadlocks and ensure the safety constraints while fulfilling a given mission by building a new PN model that accentuates the transitions that are allowed to be fired. In \cite{lacerda2019petri} is mentioned that their PN supervised control model aims to return a new composition based on a deterministic B\"uchi automaton of the LTL mission, to check the admissibility. In other words, the authors ensure that the supervisor PN does not disable events that cannot be controlled. Although this approach is suitable for formal verification that guarantees the satisfaction of the mission, the state-space explosion represents a downside that has not been fully mitigated throughout research so far.

Another perspective includes the coordination of multiple PNs in a structured manner, expressed, for example, by the NwN paradigm. This high-level abstraction guarantees a hierarchical architecture of the models, suitable for modeling both the local (\emph{Object nets}) and the global (\emph{System net}) view of a system. The object nets are represented as tokens in the system net. The object-oriented methodology suitable for high-level representations \cite{valk2003object} introduces different types of mobility among nets \cite{kohler2003modelling}, which express synchronous and asynchronous actions.

Several application examples of NwN paradigm include modeling web service coordination \cite{alvarez2005approaching}, smart houses \cite{kissoum2012modeling}, modeling the epigenetic regulation process at the cell level \cite{bardini2016using}, and simulating antibiotic resistance at the microbiota level \cite{bardini2017using}. Other works focused on self-development tools such as \emph{Renew} \cite{kummer2000renew} and \emph{Modular Model Checker} (MoMoC) \cite{willrodt2020modular}, or encoding specifications in \emph{Maude language} \cite{capra2023encoding}. 

When writing this paper, the total number of papers that involve the ``Nets-within-Nets paradigm'' is 64, using the Web of Science database. Out of this number, there are only a few papers that address the problem of predicting trajectory for single or multiple robots \cite{kohler2003modelling}. Some papers use the object-oriented or hierarchical idea of PNs, without referring to it as the NwN paradigm \cite{figat2019methodology}. Our work introduces a top-down framework with formal definitions under the NwN paradigm, suitable for heterogeneous robotic teams that ensure a global specification. In this account, the team's mission is also modeled by an object net, interacting further with nets allocated to the robots through a synchronization function that secures spatial capacity constraints by design (the maximum number of robots in an environmental region). Based on our knowledge, our proposal represents a novelty in the field of robotic planning, with the following \emph{contributions}: 

\begin{itemize}
    \item Proposing a novel framework called the \emph{High-Level robot team Petri Net} (HLrtPN) system for motion planning of heterogeneous robotic systems that ensures a global mission. For this purpose, a synchronization function (\emph{Global Enabling Function}) between the nets is designed, having the role of verifying and acting on a set of logical Boolean formulas to ensure that the specification requirements are met;
    \item Describing the step-by-step implementation of the framework in Renew \cite{kummer2000renew} and making it accessible on web \cite{NwNsite}. The illustrative examples of this implementation showcasing the modeling of LTL formulas for heterogeneous robotic teams strengthen our framework's substantial potential in robotic planning;
    \item Assessing the proposed solution through numerical simulation in Renew and comparing it with various DES representations, considering two case studies, one of which includes a real-life futuristic scenario for a robotic team evolving in a hospital. 
\end{itemize}

Currently, the framework HLrtPN facilitates an easier modeling approach in the robotic field, with the drawback that the planning strategy leading to a feasible solution requires the exploration of multiple transitions for complex systems, which can be time-consuming. Nevertheless, we consider that the potential of HLrtPN materializes further exploitation in search of viable trajectories of a heterogeneous robotic team ensuring a global mission.

\section{Background notions}\label{sec:background}

To establish a better understanding of the proposed framework, a set of fundamental concepts is being detailed in this section, e.g., Linear Temporal Logic, Nets-within-Nets paradigm, to form the necessary background for most of the readers. 

\subsection{Linear Temporal Logic}

As mentioned earlier, our goal is to provide planning solutions for multi-robot systems that must satisfy a global mission specified in Linear Temporal Logic (LTL). LTL is a formal language capable of expressing sequential and synchronous actions through a set of \textit{temporal operators}, such as: until ${\cal U}$, eventually $\diamondsuit$, always $\square$, and next $\bigcirc$ \cite{baier2008principles, Clarke99}. These operators are recursively defined over a set of atomic propositions $\mathcal{B}$. In addition, LTL makes use of Boolean operators such as \textit{conjunction} $\wedge$, \textit{implication} $\Rightarrow$, and \textit{equivalence} $\Leftrightarrow$, which can be derived from negation $\neg$ and disjunction $\vee$.  

Although the full class of LTL includes the next operator $\bigcirc$, this operator is not suitable in the context of discrete event systems that model continuous robotic dynamics \cite{Belta-RAM07, baier2008principles}. For this reason, the specifications considered in this paper are restricted to a relevant subclass of LTL formulas.  

LTL specifications can be represented as discrete event systems, most commonly through Streett or Rabin automata \cite{esparza2018one}, or through B\"uchi automata \cite{GastinOddoux2001}. In this work, the global LTL mission is translated into a non-deterministic B\"uchi automaton.  

\begin{definition}\label{def:Buchi}
The B\"{u}chi automaton (BA) corresponding to an LTL formula over the set $\mathcal{B}$ has the structure $B=\left(S,S_0,\Sigma_B,\rightarrow_B,F\right)$, where: $S$ is a finite set of states; $S_0\subseteq S$ is the (singleton) set of initial states; $\Sigma_B$ is the finite set of inputs; $\rightarrow_B\subseteq S\times\Sigma_B\times S$ is the transition relation; and $F\subseteq S$ is the set of accepting (final) states. \hfill $\blacksquare$
\end{definition}

Let $\pi(s_i,s_j)$ denote the set of all inputs enabling a transition from $s_i$ to $s_j$, expressed as a Boolean formula over $\mathcal{B}$ in Disjunctive Normal Form (DNF). An \textit{infinite accepted run} in $B$ drives the automaton from an initial state to an accepting state through (i) a \textit{prefix}, and (ii) a \textit{suffix}, which is a path returning to the same accepting state reached in the prefix. The accepted run can thus be expressed as \textit{prefix, suffix, suffix, $\ldots$} \cite{Wolper83}.  

In this paper, we focus on planning solutions for multi-robot systems under \textit{co-safe} LTL formulas. Unlike general LTL formulas, which may require infinite traces to be satisfied, co-safe LTL formulas can be satisfied by a finite execution.

\subsection{Nets-within-Nets Paradigm}

The planning solution proposed in this paper builds on the Nets-within-Nets paradigm \cite{jensen2012high, valk2003object, kohler2003modelling}. The distinctive feature of this family of high-level Petri nets is that each token can encapsulate information, such as the state of another process. In this sense, a token is represented as a Petri net, referred to as an \textit{Object Net}. The interactions among these nets are captured in a higher-level \textit{System Net}, which provides a global view of the overall system.  

This hierarchical modeling framework enables the representation of three types of behaviors, determined by transition labels. In this paper, we explicitly detail the labeling concept and its role in capturing robot motion while ensuring satisfaction of the mission. The interactions between system and object nets are as follows \cite{valk2003object}:  

- \textit{Transport}: the movement of an object net within the system net, where only the system marking is updated, while the object net retains its marking.  

- \textit{Autonomous transition}: an update occurs only in the marking of the object net, while the system net remains unchanged.  

- \textit{Interaction}: a synchronous update of both the system and object nets, triggered by simultaneous firing of transitions in each net.  

To clarify these behaviors, examples from the manufacturing domain can be provided. The \textit{transport} action corresponds to updating the global system state, such as moving a robot from one location to another, while preserving its local state (e.g., gripper closed). An \textit{autonomous transition} updates only a local state (e.g., changing from gripper closed to gripper open) while the global state (e.g., robot position) remains unchanged. Finally, an \textit{interaction} simultaneously updates both global and local states, for example, when a robot moves through the environment while opening its gripper.  

\textbf{Remark.} Typically, there is a single system net that encodes the global state of the system. In contrast, each subsystem is represented by its own object net, which can differ in structure. For instance, two distinct robots may be modeled by two different object net representations.  

In this work, we consider two types of object nets. The first, denoted \textit{SpecOPN}, models the LTL mission. Specifically, the LTL mission is first represented as a B\"uchi automaton and then translated into a Petri net following our algorithm in \cite{hustiu2024multi}. The second type, denoted \textit{RobotOPN}, captures the motion of each category of robot in the heterogeneous team. This model is related to the Quotient Robot Motion Petri Net introduced in \cite{hustiu2024multi}, but here we extend it by adopting a labeled representation of transitions as in \cite{cabasino2011discrete}.  

For both object nets, we provide formal definitions along with illustrative examples to facilitate understanding.

\section{Problem formulation}\label{sec:pbstat}

\textbf{\textit{Problem:}} This paper addresses the task allocation and planning problem for a heterogeneous multi-robot system evolving in a known environment including a set of regions of interest. The team should ensure a global mission given as a co-safe LTL specification, imposing spatial constraints of visiting/avoiding the regions, and temporal constraints requiring sequencing and synchronization.

The solution space of the problem is subject to a proposed framework under the Nets-within-Nets formalism connecting Petri net representations under a hierarchical structure. A high-level Petri net denoted \textit{system net} provides a global view of the multi-robot system with respect to a global given mission, both of them being represented by a low-level Petri net denoted \textit{object nets}. The planning solution is determined via a simulation-based approach.

An explanation of the proposed method consists of modeling the allowed movements of the heterogeneous team as a set of PNs (one assigned to each different type of robot, as sketched in Fig.~\ref{fig:NwNEx}-iii), also specifying the mission in the same formalism (as depicted through the PN in Fig.~\ref{fig:NwNEx}-i). These models will be implemented at the same hierarchical level as \emph{Object nets}. The upper-level PN, denoted \emph{System net}, considers object nets as tokens (Fig.~\ref{fig:NwNEx}-ii). Firing the transitions of the system net will impose synchronization between the object nets (of the robots and the mission).  An intuitive explanation based on Fig.~\ref{fig:NwNEx} containing the relation between the system and object nets for a multi-robot system is presented in Section \ref{sec:intuitive}.


The prerequisites of the problem formulation include the following assumptions:
\begin{itemize}
    \item Among the high-level languages to specify a mission for the robotic team, we have focused on the Linear Temporal Logic formalism, due to a provided algorithm from our previous work in \cite{hustiu2024multi}, which allows us to model the B\"uchi automaton for an LTL formula into a Petri net model. Additional details are provided in Section \ref{sec:specnet}. 
    \item The high-level planning of mobile robots returned by the proposed method is composed of a sequencing of motion plans which a low-level controller of the robots can enforce. Thus, robots are not restricted to any particular category, encapsulating various dynamic characteristics.
\end{itemize}

\subsection{Notations}

Consider a two-dimensional workspace $E$ partitioned into a finite number $w$ of disjoint areas (denoted as cells). Let $\mathcal{P} = \{P_1, P_2, \dots, P_w\}$ be the collection of cells whose union is $E$. Additionally, let $\mathcal{Y} = \{y_1, y_2, \ldots, y_q\}$ be a set of $q$ labels, also called regions of interest (ROI). The elements of $\mathcal{Y}$ are used to label the set of non-overlapping regions of the environment through a function $h: \mathcal{P} \to 2^\mathcal{Y}$ under the assumption that the label $y_q \in \mathcal{Y}$ corresponds to the free space and can be assigned to only one cell in $\mathcal{P}$. Moreover, if $y_q$ is assigned to region $P_j$, then no other label can be assigned to $P_j$. Specifically, if $y_q \in h(P_j)$, then $h(P_j) = \{y_q\}$. 

In the workspace $E$ exists a team of heterogeneous robots (agents) $\mathcal{R} = \{r_1, r_2, \ldots, r_n \}$. Initially, all $n$ robots are placed in the free-space region (the one labeled with $y_q$). At any moment, a robot should physically be placed in one cell $P_j \in \mathcal{P}$. A pre-established set of spatial constraints specifies the regions in which the robot can or cannot move. This behavior is observed in real life; for example, in a manufacturing system, one robot is allowed only to move in restricted areas for pick-and-place operations without colliding with other robots, while another robot has more freedom in its motion for monitoring the entire workspace.

Let $\mathcal{B} = \{b_1, b_2, \dots, b_q\}$ be a set of atomic propositions, where each label $y_i$ corresponds to an atomic proposition $b_i$. For any subset $A \in 2^{\mathcal{Y}}$, let us define the characteristic conjunction formula of $A$ as $A_{\wedge} = \bigwedge \{b_i \in \mathcal{B} \mid y_i \in A\}$. For example, for $A = \{y_1, y_2, y_3\}$, $A_{\wedge} = b_1 \wedge b_2 \wedge b_3$. Using the set of atomic propositions $\mathcal{B}$, the mission for the team of robots can be specified as an LTL formula \cite{Belta-RAM07}.

\subsection{Intuitive support of the proposed approach}\label{sec:intuitive}

One of the goals of this work is to provide a Nets-within-Nets (NwN) model for a heterogeneous robotic team to achieve a planning solution through simulation. The NwN model is called the \emph{High-Level Robot Team Petri Net} (HLrtPN) system, and it comprises (i) a set of object nets modeling robots (\emph{Robotic Object Petri Nets} (RobotOPNs)) and one object net modeling the mission (\emph{Specification Object Petri Net} (SpecOPN)); and (ii) one system net where each token corresponds to an object net. The \emph{system net} governs the evolution of the system. When transitions of the RobotOPNs are fired, they must fire synchronously with a transition in the system net. Additionally, when transitions in the RobotOPN are fired, the robots move between regions, updating the values of set $\mathcal{B}$. Consequently, the transition fired in the system net synchronizes with the firing of a transition in SpecOPN. The overall synchronization of the transitions in the system and object nets is ensured by the synchronization function \textit{GEF}.

\begin{figure*}
\centering
\includegraphics[width=\textwidth]{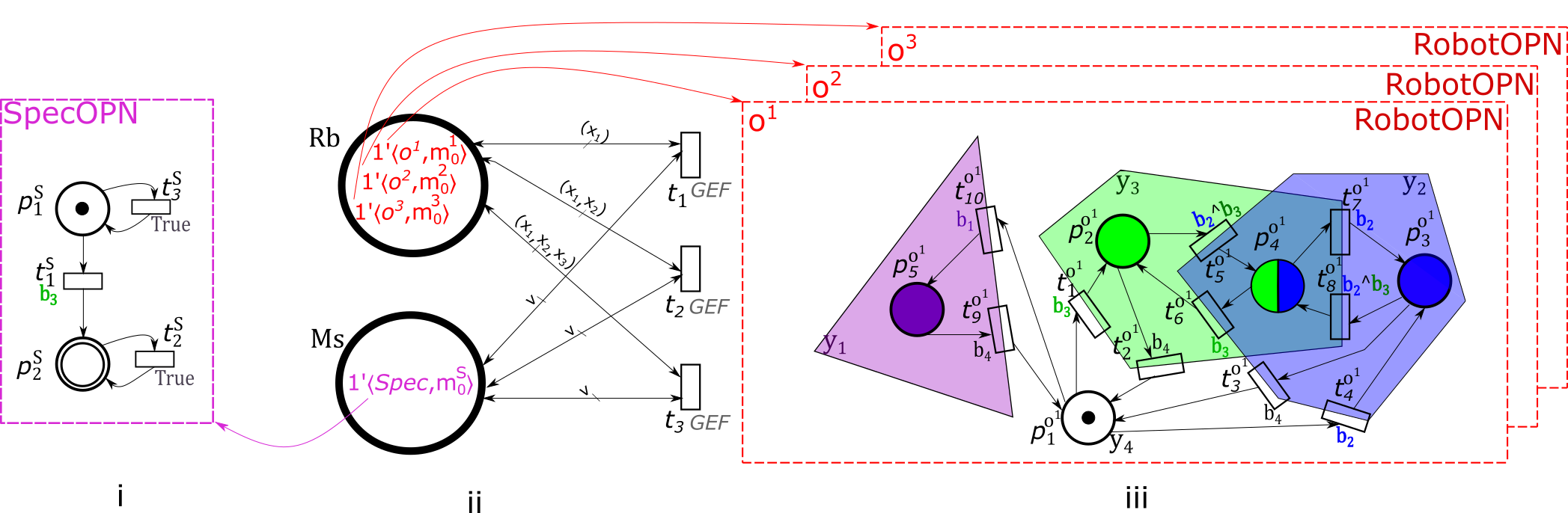}    
\caption{ Illustrative example supporting the intuitive explanation of the Nets-within-nets formalism, portraying the (ii) System net  including 2 types of object nets: (i) Specification Object Petri net and (iii) Robotic Object Petri net}  
\label{fig:NwNEx}
\end{figure*}

\begin{figure}[]
\begin{center}
\includegraphics[width=12cm]{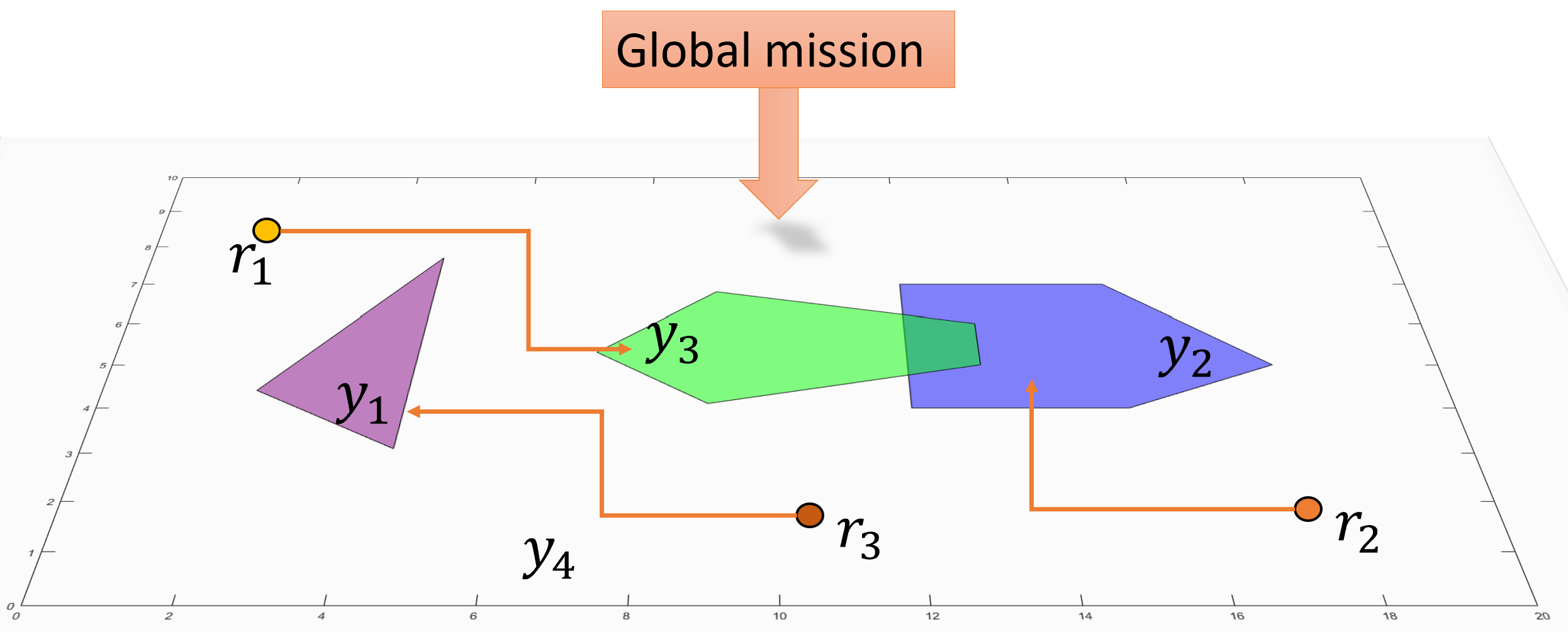}    
\caption{Example of a partitioned environment into 5 cells in set $\mathcal{P}$ including 4 ROIs ($y_1$ - purple, $y_2$ - blue, $y_3$ - green, and the free space $y_4$ - white) and 3 robots initially placed in $y_4$. The orange line represents the robotic trajectories for the mission $\varphi = \diamondsuit b_1 \wedge \diamondsuit b_2 \wedge \diamondsuit b_3 \wedge  \left(\neg b_1~\mathcal{U}~b_3\right)$ 
 (meaning to visit $y_1, y_2, y_3$, with $y_3$ before $y_1$)} 
\label{fig:env}
\end{center}
\end{figure}

Let us consider the environment in Fig.~\ref{fig:env} that can be partitioned into 5 cells $\mathcal{P}=\{P_1, \ldots, P_5\}$ where $P_1, P_2, P_3, P_4$, respectively $P_5$ are associated with the free space (white region), green region minus the intersection with blue region; the blue region minus the intersection with the green region; the intersection of the blue and green region, and the purple region. The set of labels is $y_1$ for purple region, $y_2$ for blue region, $y_3$ for green region, and $y_4$ for free space (white). Therefore, the labeling function $h$ is defined as follows: $h(P_1)=\{y_4\}$, $h(P_2)=\{y_3\}$, $h(P_3)=\{y_2\}$, $h(P_4)=\{y_2, y_3\}$ ($h(P_4)_\wedge=b_2 \wedge b_3$) and $h(P_5)=\{y_1\}$. 

The HLrtPN is illustrated in Figs. \ref{fig:NwNEx}. Fig. \ref{fig:NwNEx}-ii displays the system net with two places: $Rb$, containing three tokens, each corresponding to a RobotOPN of a robot, and $Ms$, with one token corresponding to the SpecOPN. In the system net, three transitions are included, related to the firing of a specific number of robots. Particularly, firing transition $t_1$ models the motion of one robot, any of the types of robots present in the robotic team. Similarly, firing $t_2$ models the movement of any two robots from the robotic team, while $t_3$ models the movement of the entire robotic team formed out of three robots. 

\textbf{Remark 1.} Formally, to differentiate between the object nets, we have defined the following notations such as superscripts: ``$S$'' for SpecOPN, respectively ``$o^i$'' for RobotOPN modeling the robot $r_i$, for all the components as part of these nets' definitions (Section \ref{sec:objectnet}).

Fig.\ref{fig:NwNEx}-i shows the SpecOPN for the simple formula $\varphi = \diamondsuit b_3$. The mission is assumed to be accomplished when $p_2^S$ has a token. Notice that this is possible  by firing the transition $t_1^S$ attached to the labeling function value $b_3$. The function is evaluated at $True$ when a robot enters a region labeled $y_3$ (regions $P_2$ or $P_4$).

Figs. \ref{fig:NwNEx}-iii shows one RobotOPN denoted as $o^1$. The other two are identical if the other two robots are identical. In the RobotOPN of robot $r_j$, each place models a cell from $\mathcal{P}$; specifically, for each $P_i \in \mathcal{P}$, a place $p_i^{o^j}$ is defined, where $o^j$ models the robot $r_j$. Initially, all robots are in the free space, with the initial marking of $o^j, j = \overline{1,3}$, being a token in $p_1^{o^j}$. Since the label of $P_1$ is $h(P_1)=\{y_4\}$ (the region modeled by $p_1^{o^j}$), the atomic proposition $b_4$ is evaluated as $True$ in this state. If robot $r_1$ leaves $P_1$ and enters $P_2$, transition $t_1^{o^1}$ should fire, and the atomic proposition $b_3$ becomes $True$ since $h(P_2)=\{y_3\}$. Notice that the transitions in RobotOPN are labeled with the atomic propositions evaluated as $True$. The movement of a robot from one cell to another updates the truth value of at least one atomic proposition. Adjacent cells have different labels but our modeling methodology can handle more general cases. 

If we return to the firing of $t_1^{o^1}$, assuming that the other robots are not moving, then $t_1$ from the system net should fire synchronously with $t_1^{o^1}$. In the system net, transition $t_1, t_2$, respectively $t_3$ models the movement of one, two, respectively three robot(s). Moreover, the defined \textit{GEF} ensures the synchronous firing of these two transitions with one transition from the SpecOPN, where both transitions are enabled and the logical functions assigned to them are also true ($t_3^S$ is labeled with $True$ while $t_1^S$ is labeled with $b_3$, which will become $True$ as the robot enters the green region). If transition $t_1^S$ fires, $p_2^S$ will have a token and the mission will be fulfilled.

\textbf{Remark 2.} The simulation results highlight the benefits of the current method when compared with other three DES approaches \cite{ding2011automatic, hustiu2024multi, Kloetzer2020-ey}, which were considered in Section \ref{sec:results} during the analysis results procedure and explained in Section \ref{sec:homogteam}. The last two DES methods are restricted to homogeneous robotic teams, while the first one leads to an exponential state explosion. Thus, in our current work, we have explored the versatility of the Nets-within-Nets formalism towards the representation of heterogeneous robotic teams. When a heterogeneous robot is added to the team, it is necessary to add a RobotOPN model only, allowing for easier handling of the entire framework in comparison with traditional approaches based on transition systems.

Although similar problems are studied in the literature \cite{mahulea2020path} (but mainly for identical agents), here we are concerned with designing a different formalism that allows us to combine the motion of robots with the given mission in the same model. The framework is validated through numerical experiments, based on simulations executed in dedicated software tools. Thus, the current method obtains a sub-optimal solution, rather than searching for an optimal solution by either exploring the reachability graph of various models or solving complex optimization problems. The proposed framework has the advantage of tailoring links between robots and specific tasks and has the potential to tackle complex scenarios with time analysis by including time analysis mission models specific to another formalism, as mentioned at the end of Section \ref{sec:results}.

\section{Object Petri net systems}\label{sec:objectnet}

The dynamic of the heterogeneous robotic team is modeled by a set of object nets \emph{Robotic Object Petri net} (RobotOPN), one assigned to each type of robot based on their spatial capabilities (allowed ROIs to reach). Respectively, another object net \emph{Specification Object Petri net} (SpecOPN) models the requirements of the global mission, which the team should ensure. The following subsections formally define these nets.

\subsection{Specification Object Petri net}\label{sec:specnet}

\begin{definition}[SpecOPN] \label{def:SysPN} A  \emph{Specification Object Petri net} represented by a tuple $Spec = \langle P, P_f, T, F, \lambda\rangle$, where: $P$ and $T$ are the disjoint finite set of places and transitions, $P_f\subseteq P$ is the set of final places, $F \subseteq (P \times T) \cup (T \times P)$ is the set of arcs. The transition labeling function $\lambda_{\wedge}(t) \equiv t_{\wedge}$ assigns to each transition $t \in T$ a Boolean formula defined by using the atomic propositions $\mathcal{B}$ and their negations. \hfill $\blacksquare$
\end{definition}

 The current paper considers SpecOPN, the translated PN model obtained from the B\"uchi automaton (see Algorithm 1 from \cite{hustiu2024multi} for the formal definition). The main idea is to associate each state with a place in the PN representation, while the transitions account for the transitions considered in the automaton. In particular, the proposed translation from the B\"uchi automaton adapts any disjunctive Boolean formula from the automaton $b_i \vee b_j$ to a conjunctive Boolean formula, thus returning two transitions with their labeling functions $\lambda_{\wedge}(t_x) = b_i$ and $\lambda_{\wedge}(t_z) = b_j$. The translated PN is a \emph{state machine} PN{\footnote{A state machine is a PN in which any transition has only one input and one output place.}}, and at any time, only one place is marked.

Note that the referred algorithm from \cite{hustiu2024multi} adds a set of virtual transitions for final states needed for the planning strategy, an approach that is not necessary in the current work. In addition, although the translation from a B\"uchi automaton into a B\"uchi Petri net accounts both co-safe and non-co-safe LTL, the current work considers only co-safe LTL missions since it can be ensured by a finite trace. Thus, the satisfaction of the formula is given by a final marking in the PN model, related to a final state of the automata. Since the non-co-safe LTL formulae might require an infinite trace to ensure the mission, continuous monitoring over the mission state was not considered for this work.

A marking is a $|P|$-sized $\{0,1\}$-valued vector, while a SpecOPN system is a pair $\langle Spec, \b{m}_0\rangle$ where $\b{m}_0$ is the initial marking. The specification is fulfilled when SpecOPN reaches a marking with a token in a place from $P_f$, by firing a sequence of enabled transitions. A transition $t \in T$ in the SpecOPN system is enabled at a marking $\b{m}$ when two conditions are met: (i) $\b{m}[\preset{t}] = 1$ \footnote{$^{\bullet}t$ and $t^{\bullet}$ are the input and output places of the transition $t \in T$ - singletons since the SpecOPN is a state machine.} and (ii) the Boolean formula $t_{ \wedge}$ is \emph{True}. Informally, condition (i) is the enabling condition, while (ii) means that the movement of robots with respect to the set of regions $\mathcal{Y}$ implies the firing of a transition in SpecOPN by changing the truth value of $t_{ \wedge}$. In Fig. \ref{fig:NwNEx}-i the final place is $p_2^S$, which will be marked only when $t_1^S$ fires. This is possible when a robot enters a region labeled by $y_3$.

\subsection{Robotic Object Petri net}\label{sec:robobjnet}

In this paper, we assume that RobotOPN is a state machine PN which can be considered as an \emph{labeled Petri net} \cite{cabasino2011discrete} by the addition of a labeling function over the set of transitions and places.

\begin{definition}[RobotOPN]\label{def:opn} Given a set of cells $\mathcal{P}$ modeling the workspace of the robotic team, let the \emph{Robotic Object Petri net} be the model of a robot, expressed by the tuple $o = \langle  P, T, F, h, \lambda, \gamma \rangle$: 
\begin{itemize}
    \item $P$ is the finite set of places, bijective with set $\mathcal{P}$. Each place is associated with an element $P_i \in \mathcal{P}$ in which the robot is allowed to enter;
    \item $T$ is the finite set of transitions. A transition $t_{ij} \in T$ is added between two places $p_i, p_j \in P$ only if the robot can move from any position in cell $P_i$ to cell $P_j$ in the partitioned workspace $E$;
    \item $F \subseteq (P \times T) \cup (T \times P)$ is the set of arcs. If $t_{ij}$ is the transition modeling the movement from $p_i$ to $p_j$, then $(p_i,t_{ij}) \in F$ and $(t_{ij},p_j) \in F$; 
    \item $h_{\wedge}$ is the labeling function of places $p \in P$, defined in the previous section and associating to each place a Boolean formula over the set of propositions $\mathcal{B}$; 
    \item $\lambda_{\wedge}$ is the Boolean labeling function of transitions $t \in T$, such that $\lambda_{\wedge}(t_i) = h\left(\postset{t_i}\right)_{\wedge}$;
    \item $\gamma : P \to \mathcal{P}$ is the associating function. If place $p_i \in P$ is associated with $P_i \in \mathcal{P}$, then $\gamma(p_i)=P_i$.\hfill $\blacksquare$
    \end{itemize}

\end{definition}

The marking of the RobotOPN is a vector $\b{m} \in \{0,1\}^{|P|}$. Initial marking is denoted $\b{m}_0$ such that $\b{m}_0[p_i] = 1$ if the robot is initially in $p_i$, and $\b{m}_0[p_j] = 0$ for the rest of the places $p_j \in P \setminus \{p_i\}$. A RobotOPN system is a pair $\langle o, \b{m}_0\rangle$.

 The association function $\gamma$ considers a one-to-one mapping between the set of cells $ \mathcal{P}$ and the set of places $P$, since the workspace where the robots evolve does not include regions that should be avoided by the entire multi-robot system. 

A heterogeneous robotic system incorporates the dynamics of several types of robots. We are interested in the differences concerning their space constraints (ROIs that can be reached). Each type of robot is modeled as a RobotOPN, including these differences in terms of topology and labels.

\section{High-Level robot team Petri net and GEF}\label{sec:systemnet}

\subsection{High-Level robot team Petri net}
This section introduces our proposed model denoted \emph{High-Level robot team Petri net} which encapsulates the ability to provide a global view, by enabling synchronizations between the system net and the previously defined object nets.

\begin{definition}\label{def:NwN}
A \emph{High-Level robot team Petri net} (HLrtPN) is a tuple $\N = \langle \bar{P}, \bar{T}, \mathcal{O}, \mathcal{S}, Vars, \bar{F}, W,  \mu_{cap}\rangle$, where:
\begin{itemize}
    \item $\bar{P} = \left\{Rb,Ms\right\}$ is the set of places, with $Rb$ and $Ms$ being the robot and mission places;
    \item $\bar{T}=\{t_1, t_2, \ldots, t_s\}$ is the set of transitions;
    \item $\mathcal{O} = \{\langle o^1, \b{m}_0^1\rangle, \langle o^2, \b{m}_0^2\rangle, \ldots, \langle o^n, \b{m}_0^n\rangle\}$ is a set of $n$ RobotOPN systems, one for each robot;
    \item $\mathcal{S} = \langle Spec, \b{m}_0^S \rangle$ is a SpecOPN system;
    \item $Vars=\{v, x_1, x_2,  \ldots,x_n\}$ is a set of \emph{variables};
    \item $\bar{F}$ is the set of arcs: $\bar{F}=\bigcup_{t \in \bar{T},p \in \bar{P}}\{(p,t),(t,p)\}$;
    \item $W$ is the inscription function assigning to each arc a set of variables from $Vars$ such that for every $t_i \in \bar{T}$, $W(Rb,t_i)= W(t_i,Rb) = (x_1, x_2, \ldots, x_i)$, $W(Ms,t_i) = W(t_i,Ms)= v$;
    \item $\mu_{cap}\in Bag(\mathcal{P})$ 
    is the \emph{capacity multi-set}, with $\mu_{cap}[P_i]>0, \forall i\in\{1,\ldots,w\}$ and $\mu_{cap}[P_j]\geq n$, if $h(P_j) = y_q$. \hfill $\blacksquare$
    
    
\end{itemize} 
\end{definition}

 The entire tuple $\N = \langle \bar{P}, \bar{T}, \mathcal{O}, \mathcal{S}, Vars, \bar{F}, W,  \mu_{cap}\rangle$ expresses the totality of elements as part of the proposed framework \textit{HLrtPN}. In other words, this tuple includes the notions of \textit{system net} $\langle \bar{P}, \bar{T}, \bar{F}, Vars, W\rangle$ (Fig. \ref{fig:NwNEx}-ii.), the \textit{specification net} with the initial marking $\mathcal{S}$ (Fig. \ref{fig:NwNEx}-i.), and the \textit{robotic object Petri nets} with the initial markings $\mathcal{O}$ (Fig. \ref{fig:NwNEx}-iii.). In addition, the connection between these nets, as well as introducing the capacity concept, is encapsulated through notations $ W, \mu_{cap}$.

In case of the system net, the transitions are connected via bidirectional arcs, where $t_i$ synchronizes $i$ robots according to the specification, with $i = \overline{1, s}$. The current work considers co-safe LTL missions, thus $s \leq n$. The firing of a transition manipulates the object nets by using variables, e.g., $x_1$ is bound to a RobotOPN, $v$ for the SpecOPN. Although the state of an object net is modified when a transition in the HLrtPN system is fired, by using the reference semantics as in \cite{valk2003object}, a token is a reference to an object net, the same variable being used for both directions: to/from places and transitions. 

Each cell $P_i \in \mathcal{P}$ has a given number of space units, its \emph{capacity} denoted with multi-set (also known as \textit{bag}) $\mu_{cap}[P_i]$. Specifically, it represents the maximum number of robots that can exist simultaneously in cell $P_i$. In this work, we assume that each robot has the same constant capacity of one unit. Therefore, $\forall P_i \in \mathcal{P}$ has a strictly positive capacity, considering that the free space modeled by a cell $P_j \in \mathcal{P}$ with $h(P_j) = \{y_q\}$ can accommodate the whole team (last bullet of Def. \ref{def:NwN}).


A HLrtPN system is a tuple $\langle \N, m, \mu_{occ}\rangle$ where $\N$ is a HLrtPN as Def. \ref{def:NwN}, $m$ is the marking associating a multi-set to each place in $\bar{P}$. Notice that the marking is represented as a bag $\mu$ assigning a non-negative integer number (coefficient), as it is seen in the following. The set of all multi-sets over $U$ is denoted by $Bag(U)$, as indicated in the algebra defined in \cite{Jensen1991-ym}, containing multiple operations such as addition and comparison, among others.

The initial marking $m_0$ is 
\begin{itemize}
\item $m_0[Rb] = 1'\langle o^1, \b{m}_0^1\rangle + 1' \langle o^2, \b{m}_0^2\rangle +  \ldots + 1'\langle o^n, \b{m}_0^n\rangle$;
\item $m_0[Ms] = 1'\langle Spec, \b{m}_0^S\rangle$. 
\end{itemize}

Finally, $\mu_{occ}\in Bag(\mathcal{B})$ is the \emph{occupancy multi-set} representing the actual position of the robots with respect to $\mathcal{Y}$. At a given time, $\mu_{occ}[b_i]$ is the actual number of robots in the region $y_i$. The initial occupancy bag is $\mu_{occ_0}=\sum_{i=1}^{q-1} 0'b_i + n'b_j$ since we assume that initially, all $n$ robots are in the free space.

A transition $t\in \bar{T}$ of the HLrtPN is \emph{enabled} at a given state $\langle \b{m}, \mu_{occ} \rangle$ iff
\begin{itemize}
    \item $m[Ms]$ has a transition $t^S \in T^S$ enabled;
    \item being $W(Rb,t)=(x_1, x_2, \ldots, x_i)$, $m[Rb]$ has $i$ RobotOPN net systems $(\langle o^1, \b{m}^1\rangle, \langle o^2, \b{m}^2\rangle, \ldots, \langle o^i, \b{m}^i \rangle)$ such that each of these nets has a transition $t^{o^j}$ enabled, with $ j = \overline{1,i}$, and also $GEF(\mu_{occ}, \mu_{cap}, t^S, (t^{o^1}, t^{o^2}, \ldots t^{o^i}))=True$.
\end{itemize}

An enabled transition $t\in \bar{T}$ may fire yielding the system from $\langle \m, \mu_{occ} \rangle$ to $\langle m', \mu_{occ}' \rangle$ such that,

\begin{itemize}
    \item $m'[Ms]$ has fired transition $t^S$;
    \item at $m'[Rb]$, each $o^i$ has fired transition $t^{o^i}$;
    \item $\mu_{occ}'$ is updated based on the new position of the robots.
\end{itemize}

\subsection{The Global Enabling Function (GEF)}\label{sec:GEF}

Firing a transition $t \in \bar{T}$ implies synchronized coordination at the system level, considering both the RobotOPNs from $\mathbf{m}[Rb]$ and the specification net from place $\mathbf{m}[Ms]$. This synchronization procedure must adhere to a set of compatibility rules that reflect the current state of the system, as well as $\mu_{occ}$. To ensure compliance with these rules, the \emph{Global Enabling Function} (\emph{GEF}) serves as a guard, checking the compatibility between the state of the system with the transition conditions before allowing the synchronous transitions to occur. The \emph{GEF} is essential in ensuring that the HLrtPN transition $t \in \bar{T}$ and the corresponding transitions in RobotOPNs $(t^{o^1}, t^{o^2}, \ldots, t^{o^i})$ and SpecOPN $t^S \in T^S$ proceed without breaching any rules.

The inputs considered for \emph{Global Enabling Function} which enables the firing of a transition $t \in \bar{T}$ are part of $Bag(\mathcal{P}) \times Bag(\mathcal{B}) \times T^S \times  \left( \bigcup_{i=1}^n \prod_{j=1}^{i} T_j^{o^i} \right)$. Thus, the output value returned by \textit{GEF} is either $True$, either $False$ associated with the enabling or disabling $t \in \bar{T}$. The following variables are evaluated by \textit{GEF}: the global information: occupancy bag $\mu_{occ}$, the capacity multi-set $\mu_{cap}$, a marking-enabled transition $t^S$ in the SpecOPN, $v$, and a set of $i$ marking-enabled transitions $(t^{o^1}, t^{o^2}, \ldots t^{o^i})$ in $(x_1, x_2, \ldots x_i)$.  When a number of $i$ transitions of RobotOPN are fired, the Boolean formula associated with the label of $t^S$ is satisfied, leading to an evaluation of \textit{GEF} as $True$. Contrary, the value is evaluated as $False$. Alg. \ref{alg:synchro} presents the main idea of the \textit{GEF}.

\begin{algorithm2e}[ht]
	\caption{The Global Enabling Function (\emph{GEF})}\label{alg:synchro} 
\KwIn{$\mu_{occ}, \mu_{cap}, t^S$, enabled in SpecOPN, $(t^{o^1}, t^{o^2}, \ldots t^{o^i})$,  each enabled in its corresponding RobotOPN}
\KwOut{Is the synchronized firing of $t^S,t^{o^1}, t^{o^2}, \ldots, t^{o^i}$ feasible?}
 \KwData{$(\langle o^1, \b{m}^1\rangle, \langle o^2, \b{m}^2\rangle, \ldots, \langle o^n, \b{m}^n \rangle), \mathcal{P}$}
 
Let $\chi$ be the simulated occupancy multi-set w.r.t. $\mathcal{P}$ after firing $(t^{o^1}, t^{o^2}, \ldots t^{o^i})$\;\label{update2}
\ForAll{$P_j \in \mathcal{P}$}{ \label{init_cap}
\If{$\left(\chi[P_j] > \mu_{cap}[P_j]\right)$\tcc{See Comment 1}}{ 
return $False$}
} \label{end_log}
\uIf{ $\left( t^S_{\wedge} == True \right)$\tcc{See Comment 2}}{return $True$ \label{b2}}
\Else{
Let $\mu_{occ}'$ be the simulated update of $\mu_{occ}$ w.r.t. $\mathcal{B}$ after firing $(t^{o^1}, t^{o^2}, \ldots t^{o^i})$\;\label{update3}
\ForAll{$\left(b_j \in \mathcal{B}\right)$\label{e1}\tcc{See Comment 3}}{ \label{init_verif}
\If{$\left(b_j \in t^S_{\wedge} \land \mu_{occ}'[b_j]==0\right) \lor \left(\neg b_j \in t^S_{\wedge} \land \mu_{occ}'[b_j] \geq 1\right)$ \label{verifff}}{ 
return $False$ \label{Fals} }
} \label{e2}
return $True$ \label{Tru}
}
\end{algorithm2e}

\textit{Comment 1:} Enabling a transition $t \in \bar{T}$ is verified by simulating the firing of the corresponding $i$ transitions in RobotOPN synchronized through transition $t_i$ of the HLrtPN. This artificial process is carried out by the simulation of bag $\chi$ which is computed (line \ref{update2}). In order to compute it, transitions $(t^{o^1}, t^{o^2}, \ldots, t^{o^i})$ are fictitiously fired in the corresponding RobotOPNs (from $o^1$ to $o^i$), while in the rest of RobotOPNs (from $o^{i+1}$ to $o^n$) no transition is fired. Thus, the marked places of all RobotOPN are considered. By using the associating function $\gamma^k$ in each RobotOPN, bag $\chi$ is obtained. Next, the \emph{GEF} verifies that the firing of transitions satisfies the capacity of each $P_j \in \mathcal{P}$ (line \ref{init_cap}-\ref{end_log}).  

\textit{Comment 2:} If the capacity constraints are satisfied and the Boolean formula assigned to $t^S$ (i.e., $t^S_{\wedge}$) is evaluated to $True$, then the transitions $\langle t^S, (t^{o^1}, t^{o^2}, \ldots, t^{o^i})\rangle$ can fire synchronously without being necessary the evaluation w.r.t. robots positions. As a result, the \emph{GEF} returns $True$ (line \ref{b2}). 

\textit{Comment 3:} Otherwise, a new simulation is performed and $\mu_{occ}'$ is updated. Notice that $\chi$, respectively $\mu_{occ}'$ represent the simulated occupancy multi-sets w.r.t. $\mathcal{P}$, respectively $\mathcal{B}$. In the case of $\mu_{occ}'$, there are two additional conditions that could prevent the considered transitions to fire. These conditions are checked in lines \ref{e1}-\ref{e2} and can be described as:
\begin{itemize}
\item If an atomic proposition $b_j \in \mathcal{B}$ is part of the formula $t^S_{\wedge}$, but in the simulated occupancy state obtained by the firing of the transitions ($\mu_{occ}'$), no robot will be in $y_j$. This movement of robots does not fulfill the Boolean function assigned to the transition $t^S$ (first condition in line \ref{verifff}).

\item If a negated atomic proposition $\neg b_j$ (with $b_j \in \mathcal{B}$) is part of the $t^S_{\wedge}$ formula, but the simulated update of the occupancy $\mu_{occ}'$ after the firing of the involved transitions is such that $\mu_{occ}'[b_i] \geq 1$, this means that at least one robot would be in $y_j$ and the formula $t^S_{\wedge}$ would not be fulfilled (second condition in line \ref{verifff}).  
\end{itemize}

If at least one of the previous conditions is true, then the GEF will return $False$ (line \ref{Fals}). Otherwise, it will return $True$ (line \ref{Tru}). The firing of a transition $t$ (lines \ref{b2} and \ref{Tru}) updates the marking of the system and the multi-set $\mu_{occ}$.

\begin{example}
For an easier understanding of the GEF functionality, let us consider a small example, based on Fig. \ref{fig:NwNEx}. 
Let us focus on transition $t_2 \in \bar T$, which corresponds to the synchronized action of two robots, and check whether it is enabled in the initial state shown in the figure. We assume that the capacity of each cell is equal to 1, meaning that no more than one robot can occupy a place, i.e., $\mu_{cap}[P_i]=1, i = \overline{2,5}$, while $3 \leq \mu_{cap}[P_1]$. Since the robots are initially located in the free space, the occupancy of the places is given by $\chi[P_1] = 3, \chi[P_i] = 0, i = 2,3,4,5$. The current occupancy of the robotic team with respect to the atomic proposition $\mathcal{B}$ is expressed as $\mu_{occ}[b_4] = 3, \mu_{occ}[b_i] = 0, i = 1,2,3$. At this state, $t_{1}^S$ in the SpecOPN is enabled. Since $b_3$ labels $t_{1}^S$, observation $b_3$ must be ensured after its firing.

Let us now consider several situations related to the firing of $t_2$ in the HLrtPN net, using the SpecOPN and the RobotOPNs corresponding to the first two robots:

\begin{itemize}
\item \textit{Case 1:} $t_1^S,(t_{1}^{o^1},t_{1}^{o^2})$ are all enabled. However, their synchronized firing is not feasible: in the resulting state there would be two robots in $P_2$ ($\chi[P_2] = 2$), which violates the capacity constraints (see comment 1 in the algorithm).

\item \textit{Case 2:} $t_1^S,(t_{10}^{o^1},t_{4}^{o^2})$ are also enabled. However, their synchronized firing is prevented because the resulting state would yield $\mu'_{occ}[b_3]=0$, that violates the constraint imposed to observations by $t_1^S$  (see comment 3 in the algorithm).

\item \textit{Case 3:} $t_1^S,(t_{10}^{o^1},t_{1}^{o^2})$ are enabled. In this case, firing does not violate the capacity constraints, and the resulting marking satisfies $\mu'_{occ}[b_3]=1$, which meets the condition imposed in the SpecOPN.
\end{itemize}

 As a final remark, it is important to note that the cost of any of the statements in the loops in Algorithm~\ref{alg:synchro} is constant and then, executing the GEF function is $O(\max(|\mathcal{P}|,|\mathcal{B}|))$.

\end{example}

\textbf{Remark 3.} A planning solution for the proposed system HLrtPN is returned only if the capacity constraints of the robots over a set of cells $\mathcal{P}$ mapped to the set $\mathcal{Y}$ are considered such that the mission can be accomplished. A counterexample for which the executability of our model produces a deadlock is represented by the LTL mission $\varphi = \diamondsuit y_1$ requiring the visit of the region $y_1$ for which no robot can enter due to its size. Throughout the simulations provided in the next section, we have considered feasible missions that could be satisfied by the robotic team. Thus, we have excluded scenarios that could lead to blocked or endless simulations, as previously stated.

\section{Simulations}\label{sec:results}

The proposed method is showcased by two examples, while the simulations are conducted on a computer with $12^{th}$ Gen. Intel\textregistered Core i7-12700x20 and Ubuntu 24.04LTS operating system, with 32Gb RAM. The Renew~\cite{kummer2000renew} version is 4.1.

\textit{Case Study 1 (Section \ref{sec:Easyexample})} is expressed as the simple scenario of imposing an LTL mission for a team of three robots while focusing on illustrating the concepts formally defined in the previous sections.

\textit{Case Study 2 (Section \ref{sec:Complexexample})} addresses the versatility of the method concerning the number, respectively the type (different spatial restrictions) of robots, while also investigating the computational scalability and comparing the results with other planning strategies. This experiment considers that a robotic team with up to 8 members is evolving in a pioneering upcoming context from a real environment such as a hospital.

\subsection{RENEW Implementation}

The chosen paradigm for motion planning of a heterogeneous team of robots provides a unique perspective that relies on theoretical formalism towards the development process in Renew software tool\footnote{ Renew software tool: \url{http://www.renew.de/}} \cite{kummer2000renew}. This software is a Java-based high-level Petri net simulator suitable for modeling the Nets-within-Nets paradigm in a versatile approach. Thus, the Renew implementation of our proposed formalism characterizes one of the contributions of this paper.  

\textbf{Remark 4.} It should be mentioned that we are interested in the feasibility of the LTL formulas concerning the number of robots in the team and their spatial constraints. For example, the conducted simulations for the second case study take on a mission that can be satisfied from two to eight robots. A SpecOPN model associated with an LTL formula can be generated automatically based on the defined steps documented on our website \cite{NwNsite}. In addition, the website elaborates a more thorough explanation of the mentioned notations alongside illustrative examples, as part of our entire GitHub project.

At different runs, the tool may return different solutions for the same scenario, since there may be multiple possibilities to verify the Boolean formulas from transitions of the SpecOPN. Therefore, for each experiment, a given number of simulations are performed to assess the quality of the results. The metrics chosen for the analysis of numerical results are the following.

\begin{itemize}
    \item \textbf{(a) Model size}, as the sum of places and transitions, respectively, of all the representations assigned to the robotic team and the given specification.
    \item Average \textbf{(b) run time} to return a solution, based on all the simulations as part of one experiment, such that a given LTL mission is ensured by the robotic team. This metric excludes the time needed to build the model. 
    \item Shortest \textbf{(c) trajectory length} for the whole robotic team obtained over all simulations performed within the same experiment. The trajectory length is expressed as the total number of fired transitions in the RobotOPN models.

\end{itemize}

Notice that the current approach does not guarantee an optimal robotic plan for the chosen metrics, being influenced by the randomness of performing simulations in Renew.

\subsection{Case Study 1: Easy to follow}\label{sec:Easyexample}

The first example provides an altogether view of the defined formalism, considering the problem formulation from Sec. \ref{sec:pbstat}. We exemplify a planning strategy for a team of three robots evolving in a known workspace (Fig. \ref{fig:env}) for which a global LTL specification is given. The mission $\varphi= \diamondsuit b_1 \wedge \diamondsuit b_2 \wedge \diamondsuit b_3 \wedge  \left(\neg b_1~\mathcal{U}~b_3\right)$ implies the visit of regions of interest $y_1, y_2, y_3$, but requires that region $y_3$ be visited before $y_1$. The spatial constraints for agents impose an upper bound of two, i.e., $\mu_{cap}[P_5] = 2$, meaning that no more than two robots can be present at the same time in the cell $P_5$ modeled by $p_5^{o^{1}}$, $p_5^{o^{2}}$ and $p_5^{o^{3}}$, all labeled by $y_1$.

\begin{figure*}
    \centering
    \includegraphics[width=\textwidth]{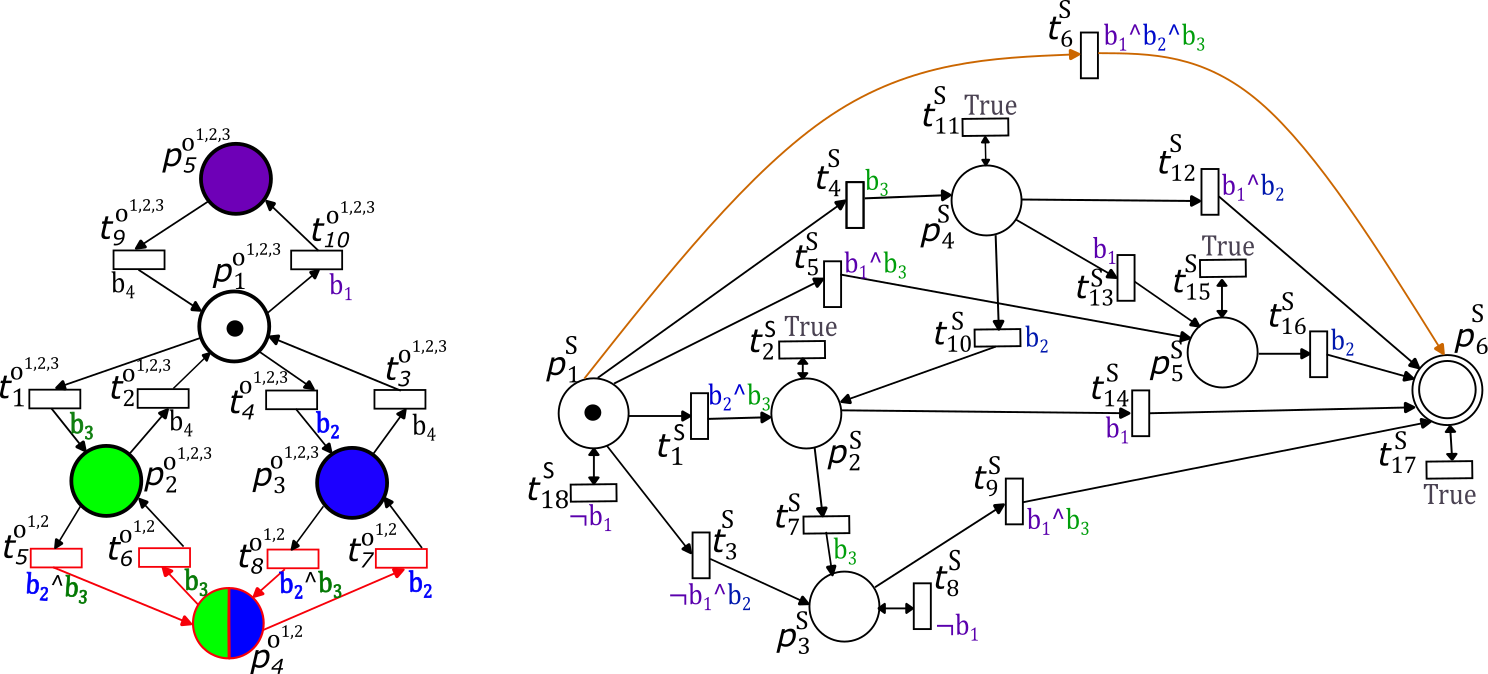}
    \caption{The object nets of the HLrtPN: (a) RobotOPN modeling the robots evolving in the environment in Fig. \ref{fig:env}: two of the robots ($r_1$ and $r_2$) can move freely in the workspace, thus being represented by the entire Petri net with all 5 places; robot $r_3$ is not allowed to enter the overlapped region between $y_2$ and $y_3$ (the Petri net representation is based only on 4 places, by removing the place, arcs and transitions with red border); (b) SpecOPN: the marked path (with dark orange) corresponds to the shortest solution for ensuring the robot mission) considering} 100 simulations according to the trajectory length of the robotic-team.
    \label{fig:ropn_spec}
\end{figure*}

Fig. \ref{fig:ropn_spec} illustrates the two types of object nets. The left side (Fig. \ref{fig:ropn_spec}-(a)), shows two different types of robots w.r.t. to their spatial capabilities: $r_1$ and $r_2$ are identical and are allowed to move freely in the workspace. In contrast, $r_3$ is not allowed to enter the overlapped part of regions of interest $y_2$ and $y_3$ (illustrated by place $p_4^{o^{1,2}}$). 

The right side (Fig. \ref{fig:ropn_spec}-(b)) shows the SpecOPN model of the LTL mission $\varphi$, resulting from algorithm \cite{hustiu2024multi}. As mentioned previously, the results in terms of robots' trajectories are returned randomly by Renew. Thus, the mean execution time per simulation (out of 100) is $\mu = 14.25$ [ms].

Let us explain the orange run from Fig. \ref{fig:ropn_spec}-(b). This run requires triggering a transition labeled $b_1 \wedge b_2 \wedge b_3$, which requires the simultaneous visit to regions $y_1$, $y_2$, and $y_3$. The mission is accomplished when place $p_6^S$ in SpecOPN contains a token, guaranteed by the synchronization function GEF based on the movements of the robots in RobotOPNs.

Initially, the robots are in free space and cannot directly enter cell $P_4$ (intersection of green and blue regions modeled by $p_4^{o^{1}}$ and $p_4^{o^{2}}$). Note that $r_3$ cannot enter this cell. Therefore, the only way to reach all three regions is for each robot to synchronously enter these regions in one step. Consequently, three robots must move, and transition $t_3$ in the system net must fire synchronously with $t_6^S$.

Furthermore, the three robots should move as follows: $t_1^{o^1}$ from $o^1$ labeled $b_3$, $t_4^{o^2}$ from $o^2$ labeled $b_2$, and $t_{10}^{o^3}$ from $o^3$ labeled $b_1$. Therefore, transition $t_3$ in the system net, which models the movement of three robots, will fire synchronously with $t_1^{o^1}$, $t_4^{o^2}$, and $t_{10}^{o^3}$ from the RobotOPNs, as well as with $t_6^S$. After these firings, a token will arrive in place $p_6^S$, fulfilling the LTL formula.

\textbf{Remark 5.} The entire representation in Renew of the HLrtPN model including the system net and the object nets can be visualized on our website \cite{NwNsite}. For an easier visualization in the Renew simulator, we have defined notations without subscripts, e.g., the formal notation of atomic propositions $\mathcal{B}$ for set $ \mathcal{Y} = \{y_1,y_2,y_3,y_4\}$ (Fig. \ref{fig:env}) is replaced in the implementation by set $\{a,b,c,w\}$, in exactly this order, with $w$ assigned to the free space $y_4$. Moreover, the symbols $\neg$ or $\wedge$ are replaced in Renew with the syntax ``$!$``, respectively ``$,$``. The $True$ value returned by the associated B\"uchi automaton of the co-safe LTL formula, is expressed in the tool with  ``1``.

\subsection{Case study 2: Assessing the flexibility of the method}\label{sec:Complexexample}

Let us consider a hospital procedure, e.g., MRI (Magnetic Resonance Imaging), suitable to scan images of the patient's body which are further used in diagnosing medical conditions or plan treatments. Due to the magnetic field generated by the machine, the computer used in the scanning process is in a different room. A radiographer usually operates the scanning process from another room. Depending on the body part that has to be monitored, the acquisition time varies, e.g., measuring the flow rates in vessels can take up to 30-40 minutes long \cite{hollingsworth2015reducing}. Due to the time-consuming process of the scanning and monitoring, the researchers are inclined to automate it, e.g., the authors of \cite{hennersperger2016towards} aim to close the gap between the current manual approach of ultrasound acquisition by using a robotic system. Since the tendency is to reach fully automated systems assisting in the medical field, many works provide different solutions approaching this aspect. One example is in\cite{li2021overview}, where various methods for the ultrasound procedure are structured based on a defined autonomy level.

\begin{figure*}[ht]
\centering
\includegraphics[width=\textwidth]{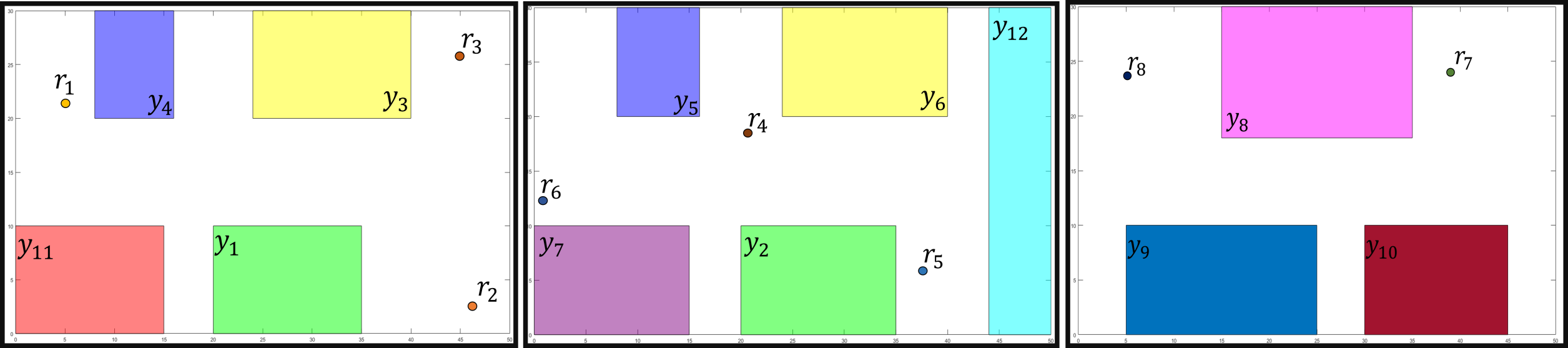}    
\caption{ Environment layout for the \textit{Case-study 2} considering a hospital scenario with three  floors and 12 rooms ($q = 12$) for a multi-robot system with 8 robots ($n = 8$).  The synchronous movements of the robots are highlighted by the same color of the rooms, considering the MRI procedure, i.e., yellow ($y_3, y_6$), green ($y_1, y_2$), blue ($y_4, y_6$). } 
\label{fig:hospital}
\end{figure*}

The need to automate this medical process among others, allows us to introduce a complex scenario suitable for motion planning of a robotic system with physical applicability in the real world. Let Fig. \ref{fig:hospital} illustrate the layout of a hospital with three floors. The hospital includes a total number of rooms of 12, denoted by the set $\mathcal{Y} = \{ y_1, y_2, y_3, y_4, y_5, y_6, y_7, y_8, y_9, y_{10}, y_{11}, y_{12}\}$, with examination rooms $y_7, y_{11}$, surgery rooms $y_8, y_{12}$, therapy rooms $y_9, y_{10}$, and MRI rooms $y_1, y_3, y_4$ which can be monitored only from rooms $y_2, y_5, y_6$.

 Multiple rooms in the hospital are reached at one point in time, as stated in the co-safe LTL mission \eqref{eq:LTLformula}. Here, a correlation between the set of regions of interest $\mathcal{Y}$ and the set of atomic proposition $\mathcal{B}$ is necessary, i.e., $b_1$ associated with $y_1$. Firstly, the patients should be first examined in one of the examination rooms, through the \textit{True} value of $b_7$ or $b_{11}$. When an MRI procedure is required afterwards, the rooms for MRI should be reached synchronously, as the patient is monitored simultaneously by a scanning robot.  In other words, pairs of atomic propositions suggesting a simultaneous reach for the MRI procedure by imposing to eventually $\diamondsuit$ visit rooms $y_1, y_2$ (expressed by $b_1, b_2$), $y_3, y_6$ (expressed by $b_3, b_6$), and $y_4, y_5$ (expressed by $b_4, b_5$). Moreover, the mission states to visit any of the therapy rooms and surgery rooms, modeling a scenario where the patient needs another procedure after the MRI. In this case, the robotic team should supply and clean these rooms through the \textit{True} value of any of the atomic propositions $b_8, b_9, b_{10}, b_{12}$.

 \begin{equation} \label{eq:LTLformula}
\begin{split}
    \varphi =  \diamondsuit (b_1 \wedge b_2) \wedge \diamondsuit (b_3 \wedge b_6)  \wedge \diamondsuit (b_4 \wedge b_5) \wedge \\
    \neg (b_1 \vee b_3 \vee b_4) \mathcal{U} (b_7 \vee b_{11}) \wedge \diamondsuit (b_8 \vee b_9 \vee b_{10} \vee b_{12})
\end{split}
\end{equation}

\subsubsection{Heterogeneous robotic team}

The robotic system includes different types of robots, based on their spatial capabilities: $r_p$ are robots carrying patients, $r_m$ have scanning abilities only for the MRI procedure, $r_{sc}$ are supplier and cleaning robots (supply with medicament and sterilize the rooms in which the patient should enter for medical operations) and $r_a$ are assistant robots having a wide range of actions, realizing the tasks of $r_m$ and $r_{sc}$. Table \ref{tb:agents_rooms} illustrates the agents' capabilities w.r.t. the spatial constraints. For example, agents $r_p$ can only enter rooms $y_1, y_3, y_4, y_7, y_{11}$ for MRI or leading the patients for examination, while agents $r_m$ have access only in rooms $y_2, y_5, y_6$ to scan the patient during the MRI procedure. 

The result analysis is executed for teams of two to eight robots, shown in Table \ref{tb:scalab}. The first columns of the table present the cardinality of each type of robot for every scenario. It is observed that the \textbf{(a) model size} directly influences the \textbf{(b) run time}. These simulations prove that the proposed framework satisfies the main objective in terms of a motion plan for a multi-agent system, considering offline planning. Thus, the \textbf{(c) trajectory length} could be shortened by introducing an optimality problem, the visualized result currently being computed through random solutions.

\textbf{Remark 6.} Generally, the proposed framework ensures solutions in which a subset of the robotic team synchronizes. This subset is a user-defined agent group bounded by the team cardinality. Particularly, the second case study for heterogeneous robotic teams generates solutions determined by a subset equal to the entire set of the robotic team (Table \ref{tb:scalab}). 

\begin{table}[!ht]
\centering 
\caption{Robots spatial capabilities w.r.t. the hospital's rooms.}\label{tb:agents_rooms}

        \begin{tabular}{|c|c|c|c|c|c|}
        \hline \hline \multirow{2}{3em}{Floor} & \multirow{2}{3em}{Rooms} & \multicolumn{4}{c|}{Robots} \\ 
        
        & & $r_p$ &  $r_m$ & $r_{sc}$ & $r_a$\\
        \hline
        \RomanNumeralCaps{1} & $y_1$ & $\times$ & & &  \\
        \hline
        \RomanNumeralCaps{2} & $y_2$ & & $\times$ & & $\times$  \\
        \hline
        \RomanNumeralCaps{1} & $y_3$ & $\times$ & & &   \\      
        \hline
        \RomanNumeralCaps{1} & $y_4$ & $\times$ & & &   \\      
        \hline
        \RomanNumeralCaps{2} & $y_5$ & & $\times$ & & $\times$  \\     
        \hline
        \RomanNumeralCaps{2} & $y_6$ & & $\times$ & & $\times$  \\
        \hline
        \RomanNumeralCaps{2} & $y_7$ & $\times$ & & &   \\ 
        \hline
        \RomanNumeralCaps{3} & $y_8$ & & & $\times$ & $\times$ \\ 
        \hline
        \RomanNumeralCaps{3} & $y_9$ & & & $\times$ & $\times$ \\ 
        \hline
        \RomanNumeralCaps{3} & $y_{10}$ & & & $\times$ & $\times$ \\   
        \hline
        \RomanNumeralCaps{1} & $y_{11}$ & $\times$ & & &   \\
        \hline
        \RomanNumeralCaps{2} & $y_{12}$ & & & $\times$ & $\times$ \\ 
        \hline
        \end{tabular}
\end{table}

\subsubsection{Homogeneous robotic team}\label{sec:homogteam}

Let $r_f$ be a full robot that is not restricted in movement and has access to all rooms. Explicitly, the robot can carry patients in the examination and MRI rooms, it has the necessary abilities to scan the patient for the MRI process, as well as being able to carry supplies and clean the therapy and surgery rooms. When the team includes only this type of robot, the team consists of identical agents. Thus, the current method is evaluated alongside other Discrete Event Systems approaches, suitable for motion planning of homogeneous teams ensuring a global LTL specification. The methods are briefly outlined below. The first two are mathematical programming based on PN approaches, while the third is based on graph-search algorithms.

\begin{itemize}
    \item[(i)] \textbf{WB \cite{hustiu2024multi}} - this work (\textit{with B\"uchi (WB)}) proposes a parallel approach between (1) a reduced Petri net model assigned to the motion of the robotic team in a partitioned workspace and (2) a B\"uchi automaton modeling the given mission under the LTL formalism, which is further translated into a B\"uchi Petri net model. The models are connected through an intermediate layer of places modeling the atomic propositions, building a Composed Petri net model. Two MILPs (Mixed Integer Linear Programming) problems compute the motion plan: using the Composed Petri net model, and projecting this solution into the full PN model of the discretized environment. This work guarantees collision-free trajectories based on imposed constrained in the optimization problems, but it cannot guarantee the completeness of the algorithm due to the projection step.
    \item[(ii)] \textbf{FB \cite{Kloetzer2020-ey}} - this method (\textit{following B\"uchi (WB)}) aims for a sequential approach of computing trajectories of the robotic team by following a path from the initial state towards one final state (known also as an accepted run) in the B\"uchi automaton. First, the PN model is assigned to the team, based on the partitioned workspace. Secondly, an optimization MILP problem is solved in the search for a sequence of markings that can generate the necessary observations that fulfill the LTL mission based on a set of feasible runs computed in the B\"uchi automaton. The approach is iterative until the team can act under the accepted run in the automaton, producing a sub-optimal solution that cannot ensure collision-free trajectories.  
    \item[(iii)] \textbf{TS \cite{ding2011automatic}} - this approach (\textit{Transition Systems (TS)}) is subject to represent the motion of each robot as a Transition System model versus the previous approaches where one single model was assigned for the entire robotic team. Robots' trajectories are computed by a graph-search-based algorithm for the composed model built from all the individual models of the robots together with the automata model of the LTL mission. 
\end{itemize}

\begin{table*}[!ht]
\centering
\caption{Comparison results for the homogeneous robotic team between the current and other DES methods from literature}\label{tb:same_rob}
  \begin{adjustbox}{width=\textwidth}
\begin{tabular}{|c|c|c|c|c|}
        \hline \hline No. of rob & Algorithm & \textbf{(a) Model size} & \textbf{(b) Run time} [s] &\textbf{(c) Trajectory length} \\ 
        \hline \hline
        \multirow{4}{1em}{2} & \textbf{HLrtPN}  & $(|P|,|T|) = (50, 322)$ & 0.8 & 33 \\
        & \textbf{WB \cite{hustiu2024multi}}& $(|P|,|T|) = (55, 288)$ & 0.86 & 14 \\
        & \textbf{FB \cite{Kloetzer2020-ey}} &  $(|P|,|T|) = (13, 26), (|N_B|, |T_B|) = (18, 108)$ & 0.97 & 14 \\
        & \textbf{TS \cite{ding2011automatic}} & $|N_n| = 3042$ & 1.51 & 14 \\
        \hline
        \multirow{4}{1em}{3} & \textbf{HLrtPN} &$(|P|,|T|) = (65, 351)$ & 0.5 & 28 \\
        & \textbf{WB \cite{hustiu2024multi}}&$(|P|,|T|) = (55, 288)$ & 1.1 & 13 \\
        & \textbf{FB \cite{Kloetzer2020-ey}}&$(|P|,|T|) = (13, 26), (|N_B|, |T_B|) = (18, 108)$ & 0.9 & 13 \\
        & \textbf{TS \cite{ding2011automatic}}&$|N_n| = 3.9 * 10^3$ & 1940.33 & 13 \\
        \hline
        \multirow{4}{1em}{4} & \textbf{HLrtPN} &$(|P|,|T|) = (80, 380)$ & 4.5 & 19 \\
        & \textbf{WB \cite{hustiu2024multi}}&$(|P|,|T|) = (55, 288)$ & 0.71 & 12 \\
        & \textbf{FB \cite{Kloetzer2020-ey}}&$(|P|,|T|) = (13, 26), (|N_B|, |T_B|) = (18, 108)$ & 0.76 & 12 \\
        & \textbf{TS \cite{ding2011automatic}}&$|N_n| = 5.1 * 10^4 $ & $\approx 3$  $days$ & $-$  \\
        \hline
        \multirow{4}{1em}{5} & \textbf{HLrtPN} &$(|P|,|T|) = (95, 409)$ & 10.9 & 17 \\
        & \textbf{WB \cite{hustiu2024multi}} &$(|P|,|T|) = (55, 288)$ & 0.74 & 11 \\
        & \textbf{FB \cite{Kloetzer2020-ey}} &$(|P|,|T|) = (13, 26), (|N_B|, |T_B|) = (18, 108)$ & 0.88 & 11 \\
        & \textbf{TS \cite{ding2011automatic}}&$|N_n| = 6.6 * 10^5 $ & $-$ & $-$  \\
        \hline
         \multirow{4}{1em}{6} & \textbf{HLrtPN} & $(|P|,|T|) = (110, 438)$ & 39.5 & 24 \\
        & \textbf{WB \cite{hustiu2024multi}} &$(|P|,|T|) = (55, 288)$ & 0.62 & 10 \\
        & \textbf{FB \cite{Kloetzer2020-ey}} &$(|P|,|T|) = (13, 26), (|N_B|, |T_B|) = (18, 108)$ & 0.88 & 10 \\
        & \textbf{TS \cite{ding2011automatic}}& $|N_n| = 8.6 * 10^6 $ & $-$ & $-$  \\
        \hline
        \multirow{4}{1em}{7} & \textbf{HLrtPN} &$(|P|,|T|) = (125, 467)$ & 133.2 & 26 \\
        & \textbf{WB \cite{hustiu2024multi}} &$(|P|,|T|) = (55, 288)$ & 0.74 & 9 \\
        & \textbf{FB \cite{Kloetzer2020-ey}}&$(|P|,|T|) = (13, 26), (|N_B|, |T_B|) = (18, 108)$ & 1.41 & 9 \\
        & \textbf{TS \cite{ding2011automatic}}&$|N_n| = 1.1 * 10^9 $ & $-$ & $-$  \\
        \hline
        \multirow{4}{1em}{8} & \textbf{HLrtPN} &$(|P|,|T|) = (140, 496)$ & 227.7 & 16 \\
        & \textbf{WB \cite{hustiu2024multi}} &$(|P|,|T|) = (55, 288)$ & 0.17 & 8 \\
        & \textbf{FB \cite{Kloetzer2020-ey}}&$(|P|,|T|) = (13, 26), (|N_B|, |T_B|) = (18, 108)$ & 1.43 & 8 \\
        & \textbf{TS \cite{ding2011automatic}} &$|N_n| = 1.4 * 10^{10}$ & $-$ & $-$  \\
        \hline
        \end{tabular}
\end{adjustbox}
\end{table*}

\textbf{Remark 7.} To maintain the consistency of the comparison procedure, all the mentioned methods, including the current one, are subject to the smallest discrete representation of the environment w.r.t. the number of partition elements, i.e., one element is associated with a single atomic proposition. Moreover, these methods are integrated into RMTool - MATLAB \cite{IPGoMaKl15}, thus making them accessible for any simulation according to the user's needs. 

As previously stated, the team's model size represents one metric taken into consideration for evaluation purposes. Thus, let us express the size of the models for each of the mentioned methods as follows: 
\begin{itemize}
    \item \textbf{WB \cite{hustiu2024multi}} - the total number of places and transitions $(|P|,|T|)$ of the defined Composed Petri net model given by the sum of the size of the Petri net model associated with the environment, the size of the B\"uchi Petri net model associated with the LTL formula, and the number of places for the intermediate layer.
    \item \textbf{FB \cite{Kloetzer2020-ey}} - the number of places and transitions of the Petri net model of the environment $(|P|,|T|)$, as well as the size of the B\"uchi automaton $(|N_B|, |T_B|)$ of the LTL formula, since both models are examined sequentially.
    \item \textbf{TS \cite{ding2011automatic}} - the total number of nodes in the product automata $|N_n| = |N_{s_r}|^n \times |N_B|$ considering the size of the transition system for each robot and the size of the B\"uchi automaton.
\end{itemize}

Notice that the first two methods have fixed sizes of models regardless of the number of robots in the team versus the last method which is strongly dependent on the size of the team, leading to a state-space explosion that is difficult to maintain for computational operations.

\begin{table*}[!ht]
\centering
        \caption{Comparison results for heterogeneous robotic team for the proposed approach}\label{tb:scalab}
\begin{adjustbox}{width=\textwidth}
\begin{tabular}{|c|c|c|c|c|c|c|c|c|}
        \hline \hline \multirow{2}{5em}{No. of rob} & \multicolumn{4}{c|}{Types of robots}  & \multirow{2}{8em}{No. of simulations} &  \multirow{2}{7em}{\textbf{(a) Model size}} & \multirow{2}{7em}{\textbf{(b) Run time} [s]} & \multirow{2}{10em}{\textbf{(c) Trajectory length}} \\ 
        & $r_p$ &  $r_m$ & $r_{sc}$ & $r_a$ & & & &  \\
        \hline \hline
        2 & 1 & & & 1 & 1000 &  $(|P|,|T|) = (42,294)$ & 0.39 & 27  \\
        \hline
        3 & 1 & 1 & 1 & & 1000 & $(|P|,|T|) = (37,293)$ & 0.24 & 29 \\
        \hline
        4 & 2 & 1 & 1 & & 1000 & $(|P|,|T|) = (44,305)$ & 1.1 & 25 \\ 
        \hline
        5 & 2 & 2 & 1 & & 1000 &  $(|P|,|T|) = (48,311)$ & 1.89 & 15 \\      
        \hline
        6 & 2 & 2 & 2 & & 1000 & $(|P|,|T|) = (54,321)$ &  10.4 & 14 \\  
        \hline
        7 & 2 & 2 & 2 & 1 & 250 & $(|P|,|T|) = (69,337)$ & 107.15 & 20 \\
        \hline
        8 & 3 & 2 & 2 & 1 & 245 & $(|P|,|T|) = (76,349)$ & 228.91 & 16 \\ 
        \hline
        \end{tabular}
    \end{adjustbox}

\end{table*}

The notation HLrtPN will refer to the proposed method, for an easier visualization in the comparison Table \ref{tb:same_rob}. Let us introduce the notation $(|P|,|T|)$ to capture the size of the entire model, where $|P|$ and $|T|$ are computed similarly, i.e., $|P| = \sum_{k = 1}^n |P^{o^k}| + |P^S| + 2$. The result represents the sum of all RobotOPN models for each robot $r_k$, the size of SpecOPN, and the size of the system net. As defined, the latter representation includes only two places $Rb, Ms$, and the number of transitions is equal to the number of robots in the team. For this scenario, the size of RobotOPN for $r_f$ is $(15,28)$ (considering one free space place for each floor of the hospital) and $(18,264)$ for the size of SpecOPN due to the automated generation from a B\"uchi automaton.

Table \ref{tb:same_rob} illustrates a comparative study between the current approach and the mentioned relevant Petri net approaches, which embody the defined performance metrics with numerical values. Note that the described methods \textbf{(i), (ii), (iii)} do not require multiple simulations for one experiment. Therefore, the \textbf{(b) run time} and \textbf{(c) trajectory length} are computed only once, without the need to compute an average metric for \textbf{(b)} or return the shortest trajectory for \textbf{(c)}. The solver used for approaches \textbf{(i), (ii), (iii)} is CPLEX for MATLAB.

The \textbf{(b) run time} for our proposed work represents the mean time for each experiment \footnote{1000 simulations for the first three cases (2-5 robots), 250 simulations for 6 robots, 85 simulations for 7 robots, and 300 simulations for 8 robots.}. As observed, the HLrtPN model tends to exhibit steeper increases in running time when more robots are added to the team, compared with \textbf{WB \cite{hustiu2024multi}, FB \cite{Kloetzer2020-ey}}, on account of the number of branches explored by the Renew simulator. The last metric \textbf{(c)} portrays the comparison of the trajectory length for the entire robotic team, the smallest value being computed for methods \textbf{WB \cite{hustiu2024multi}, FB \cite{Kloetzer2020-ey}} an account of the optimization problems. In the case of HLrtPN, we expect this metric to decrease for a higher number of simulations for one experiment.

\textbf{Remark 8.} In the Table \ref{tb:scalab}, it can be observed that the number of transitions increases with the number of robots in the team. Still, the model can be handled, and planning solutions can be computed for up to 8 robots in the team. Note that here, the increase considers only the type of robots, independently of the number of robots included in each category. In contrast, the approach \textbf{TS \cite{ding2011automatic}} leads to a complex model, since a product needs to be built. This model becomes too large to be computationally tractable for teams of more than 4 robots. 




Although there are DES methods that are more cost-effective in terms of performance for metrics \textbf{(b), (c)} for homogeneous teams, our proposed method is shown to be efficient for heterogeneous teams, due to its flexibility by design, as noted in Table \ref{tb:scalab}.

\section{Conclusions and Future Work}

In conclusion, the work addresses the problem of motion planning for a team of heterogeneous robots ensuring a global mission that includes visiting and/or avoiding regions of interest. Our proposed novel framework, denoted \emph{High-Level robot team Petri Net} (HLrtPN) system, exploits the Nets-within-Nets paradigm to model the movement and mission of the robots through a nested Petri net structure. The \emph{Global Enabling Function} ensures synchronization between the different nets. The step-by-step implementation in Renew is available in \cite{NwNsite}. The feasibility of HLrtPN is supported by simulation experiments considering a realistic hospital scenario, facilitating easier modeling of robotic systems when compared with other discrete event system approaches. 


\textbf{Limitations and future work.} Besides the series of advantages that our proposed framework introduces, such as the novelty of the Nets-within-Nets paradigm used for the first time (based on the current state of the art) in planning solutions for multi-robot systems, the flexibility towards modeling heterogeneous robotic teams ensuring a global mission while having comparable simulation results when considering other approaches from literature \cite{Kloetzer2020-ey, hustiu2024multi, ding2011automatic}, some limitation can be noted here. Four challenges are raised in terms of (i) computing the optimal solution, (ii) inferring the computational tractability scales when the number of robots increases, (iii) encapsulating time constraints under high-level Petri nets formalism, which would be suitable in the motion planning field considering real-time applications, and (iv) preventing the occurrence of spurious firing of transitions. We believe future work can be directed towards these alleviated issues, enhancing its potential.

\section{CRediT authorship contribution statement}

\textbf{Sofia Hustiu}: Conceptualization, Methodology, Formal analysis, Investigation, Writing - Original Draft;
\textbf{Joaquín Ezpeleta}: Conceptualization, Investigation, Methodology, Software;
\textbf{Cristian Mahulea}: Conceptualization, Methodology, Supervision, Writing - Review \& Editing;
\textbf{Marius Kloetzer}: Conceptualization, Methodology, Formal analysis, Writing - Review \& Editing.

\section{Declaration of competing interest}

The authors declare that they have no known competing financial interests or personal relationships that could have appeared to influence the work reported in this paper.

\section{Acknowledgment}
We acknowledge the support from projects PID2024-159284NB-I00, PID2023-148202OB-C22 and TED2021-130449B-I00 of the Spanish Ministry of Science, Innovation, and Universities, and from Grant N62909-24-1-2081 of the US Department of the Navy.

We are grateful to Eva Robillard (University of Paris-Saclay) for her assistance with the implementation during the initial stages of this work. We also extend our sincere thanks to Octavian-Cezar Pastravanu for his insightful discussions, which were instrumental in shaping the depth of this study. Finally, we acknowledge with appreciation the constructive comments and suggestions of the Editors and Reviewers, which greatly enhanced the quality of the manuscript.


\begin{thebibliography}{10}
\expandafter\ifx\csname url\endcsname\relax
  \def\url#1{\texttt{#1}}\fi
\expandafter\ifx\csname urlprefix\endcsname\relax\def\urlprefix{URL }\fi
\expandafter\ifx\csname href\endcsname\relax
  \def\href#1#2{#2} \def\path#1{#1}\fi



\bibitem{LAV06}
S.~M. LaValle, {Planning Algorithms}, Cambridge University Press, United Kingdom, 2006.
\newblock \href {https://doi.org/http://dx.doi.org/10.1017/CBO9780511546877} {\path{doi:http://dx.doi.org/10.1017/CBO9780511546877}}.

\bibitem{lindemann2017robust}
L.~Lindemann, D.~V. Dimarogonas, Robust motion planning employing {Signal Temporal Logic}, in: ACC 2017: American Control Conf., 2017, pp. 2950--2955.
\newblock \href {https://doi.org/10.23919/ACC.2017.7963399} {\path{doi:10.23919/ACC.2017.7963399}}.

\bibitem{mehdipour2020specifying}
N.~Mehdipour, C.-I. Vasile, C.~Belta, Specifying user preferences using weighted signal temporal logic, IEEE Control Systems Letters 5~(6) (2020) 2006--2011.
\newblock \href {https://doi.org/10.1109/LCSYS.2020.3047362} {\path{doi:10.1109/LCSYS.2020.3047362}}.

\bibitem{IPMaKlLe20}
C.~Mahulea, M.~Kloetzer, J.-J. Lesage, Multi-robot path planning with boolean specifications and collision avoidance, IFAC-PapersOnLine 53~(4) (2020) 101--108.
\newblock \href {https://doi.org/https://doi.org/10.1016/j.ifacol.2021.04.011} {\path{doi:https://doi.org/10.1016/j.ifacol.2021.04.011}}.

\bibitem{yu2022security}
X.~Yu, X.~Yin, S.~Li, Z.~Li, Security-preserving multi-agent coordination for complex temporal logic tasks, Control Engineering Practice 123 (2022) 105--130.
\newblock \href {https://doi.org/https://doi.org/10.1016/j.conengprac.2022.105130} {\path{doi:https://doi.org/10.1016/j.conengprac.2022.105130}}.

\bibitem{plaku2016motion}
E.~Plaku, S.~Karaman, Motion planning with temporal-logic specifications: Progress and challenges, AI communications 29~(1) (2016) 151--162.
\newblock \href {https://doi.org/10.3233/AIC-150682} {\path{doi:10.3233/AIC-150682}}.

\bibitem{tabuada2006linear}
P.~Tabuada, G.~J. Pappas, Linear time logic control of discrete-time linear systems, IEEE Transactions on Automatic Control 51~(12) (2006) 1862--1877.
\newblock \href {https://doi.org/10.1109/TAC.2006.886494} {\path{doi:10.1109/TAC.2006.886494}}.

\bibitem{cassandras2008introduction}
C.~G. Cassandras, S.~Lafortune, Introduction to discrete event systems, Springer, 2008.
\newblock \href {https://doi.org/http://dx.doi.org/10.1007/978-0-387-68612-7} {\path{doi:http://dx.doi.org/10.1007/978-0-387-68612-7}}.

\bibitem{Belta-RAM07}
C.~Belta, A.~Bicchi, M.~Egerstedt, E.~Frazzoli, E.~Klavins, G.~Pappas, {Symbolic planning and control of robot motion}, IEEE Robotics and Automation Magazine 14~(1) (2007) 61--71.
\newblock \href {https://doi.org/10.1109/MRA.2007.339624} {\path{doi:10.1109/MRA.2007.339624}}.

\bibitem{lacerda2019petri}
B.~Lacerda, P.~U. Lima, Petri net based multi-robot task coordination from temporal logic specifications, Robotics and Autonomous Systems 122 (2019) 103289.
\newblock \href {https://doi.org/https://doi.org/10.1016/j.robot.2019.103289} {\path{doi:https://doi.org/10.1016/j.robot.2019.103289}}.

\bibitem{mahulea2020path}
C.~Mahulea, M.~Kloetzer, R.~Gonz{\'a}lez, Path Planning of Cooperative Mobile Robots Using Discrete Event Models, John Wiley \& Sons, 2020.

\bibitem{ju2021hybrid}
C.~Ju, H.~I. Son, A hybrid systems-based hierarchical control architecture for heterogeneous field robot teams, IEEE Transactions on Cybernetics (2021).
\newblock \href {https://doi.org/10.1109/TCYB.2021.3133631} {\path{doi:10.1109/TCYB.2021.3133631}}.

\bibitem{jensen2012high}
K.~Jensen, G.~Rozenberg, High-level {P}etri nets: theory and application, Springer Science \& Business Media, 2012.

\bibitem{valk2003object}
R.~Valk, {Object {Petri} nets}, in: Advanced Course on {Petri} Nets, Springer, 2003, pp. 819--848.
\newblock \href {https://doi.org/http://dx.doi.org/10.1007/978-3-540-27755-2\_23} {\path{doi:http://dx.doi.org/10.1007/978-3-540-27755-2\_23}}.

\bibitem{NwNsite}
{Nets-within-Nets for Motion Planning with Renew}, \url{https://Sof16.github.io/}, accessed: 15-02-2025 (2025).

\bibitem{madridano2021trajectory}
{\'A}.~Madridano, A.~Al-Kaff, D.~Mart{\'\i}n, A.~De~La~Escalera, Trajectory planning for multi-robot systems: Methods and applications, Expert Systems with Applications 173 (2021) 114660.
\newblock \href {https://doi.org/https://doi.org/10.1016/j.eswa.2021.114660} {\path{doi:https://doi.org/10.1016/j.eswa.2021.114660}}.

\bibitem{antonyshyn2023multiple}
L.~Antonyshyn, J.~Silveira, S.~Givigi, J.~Marshall, Multiple mobile robot task and motion planning: A survey, ACM Computing Surveys 55~(10) (2023) 1--35.
\newblock \href {https://doi.org/https://doi.org/10.1145/3564696} {\path{doi:https://doi.org/10.1145/3564696}}.

\bibitem{ding2011automatic}
X.~C. Ding, M.~Kloetzer, Y.~Chen, C.~Belta, Automatic deployment of robotic teams, IEEE Robotics \& Automation Magazine 18~(3) (2011) 75--86.
\newblock \href {https://doi.org/10.1109/MRA.2011.942117} {\path{doi:10.1109/MRA.2011.942117}}.

\bibitem{kloetzer2015ltl}
M.~Kloetzer, C.~Mahulea, {LTL}-based planning in environments with probabilistic observations, IEEE Transactions on Automation Science and Engineering 12~(4) (2015) 1407--1420.
\newblock \href {https://doi.org/10.1109/TASE.2015.2454299} {\path{doi:10.1109/TASE.2015.2454299}}.

\bibitem{luo2022temporal}
X.~Luo, M.~M. Zavlanos, Temporal logic task allocation in heterogeneous multirobot systems, IEEE Transactions on Robotics 38~(6) (2022) 3602--3621.
\newblock \href {https://doi.org/10.1109/TRO.2022.3181948} {\path{doi:10.1109/TRO.2022.3181948}}.

\bibitem{ju2019modeling}
C.~Ju, H.~I. Son, Modeling and control of heterogeneous agricultural field robots based on {Ramadge--Wonham} theory, IEEE Robotics and Automation Letters 5~(1) (2019) 48--55.
\newblock \href {https://doi.org/10.1109/LRA.2019.2941178} {\path{doi:10.1109/LRA.2019.2941178}}.

\bibitem{hustiu2023extension}
I.~Hustiu, M.~Kloetzer, C.~Mahulea, Extension of a decomposition method for a global {LTL} specification, in: ETFA 2023: Int. Conf. on Emerging Technologies and Factory Automation, IEEE, 2023, pp. 1--4.
\newblock \href {https://doi.org/10.1109/ETFA54631.2023.10275529} {\path{doi:10.1109/ETFA54631.2023.10275529}}.

\bibitem{hustiu2024multi}
S.~Hustiu, C.~Mahulea, M.~Kloetzer, J.-J. Lesage, {On multi-robot path planning based on Petri net models and LTL specifications}, IEEE Transactions on Automatic Control (2024).
\newblock \href {https://doi.org/10.1109/TAC.2024.3386024} {\path{doi:10.1109/TAC.2024.3386024}}.

\bibitem{figat2022parameterised}
M.~Figat, C.~Zieli{\'n}ski, Parameterised robotic system meta-model expressed by {Hierarchical Petri nets}, Robotics and Autonomous Systems 150 (2022) 103987.
\newblock \href {https://doi.org/https://doi.org/10.1016/j.robot.2021.103987} {\path{doi:https://doi.org/10.1016/j.robot.2021.103987}}.

\bibitem{allison2022modeling}
M.~Allison, M.~Spradling, Modeling sub-team formations for heterogeneous multi-robot systems using colored {Petri}-net semantics, in: 2022 18th Int. Conf. on Distributed Computing in Sensor Systems (DCOSS), IEEE, 2022, pp. 237--243.
\newblock \href {https://doi.org/10.1109/DCOSS54816.2022.00048} {\path{doi:10.1109/DCOSS54816.2022.00048}}.

\bibitem{schillinger2021adaptive}
P.~Schillinger, S.~Garc{\'\i}a, A.~Makris, K.~Roditakis, M.~Logothetis, K.~Alevizos, W.~Ren, P.~Tajvar, P.~Pelliccione, A.~Argyros, et~al., Adaptive heterogeneous multi-robot collaboration from formal task specifications, Robotics and Autonomous Systems 145 (2021) 103866.
\newblock \href {https://doi.org/https://doi.org/10.1016/j.robot.2021.103866} {\path{doi:https://doi.org/10.1016/j.robot.2021.103866}}.

\bibitem{kohler2003modelling}
M.~K{\"o}hler, D.~Moldt, H.~R{\"o}lke, Modelling mobility and mobile agents using {Mets within nets}, in: ICATPN 2003: Applications and Theory of {Petri} Nets, Eindhoven, The Netherlands, June 23--27, Springer, 2003, pp. 121--139.

\bibitem{alvarez2005approaching}
P.~Alvarez, J.~A. Banares, J.~Ezpeleta, Approaching web service coordination and composition by means of {P}etri {N}ets. the case of the {Nets-within-nets} paradigm, in: Int. Conf. on Service-Oriented Computing, Springer, 2005, pp. 185--197.

\bibitem{kissoum2012modeling}
Y.~Kissoum, R.~Maamri, Z.~Sahnoun, Modeling smart home using the paradigm of nets within nets, in: AIMSA 2012: Artificial Intelligence: Methodology, Systems, and Applications, Varna, Bulgaria, September 12-15, 2012., Springer, 2012, pp. 286--295.
\newblock \href {https://doi.org/http://dx.doi.org/10.1007/978-3-642-33185-5\_32} {\path{doi:http://dx.doi.org/10.1007/978-3-642-33185-5\_32}}.

\bibitem{bardini2016using}
R.~Bardini, A.~Benso, S.~Di~Carlo, G.~Politano, A.~Savino, Using {Nets-within-nets} for modeling differentiating cells in the epigenetic landscape, in: IWBBIO 2016: Bioinformatics and Biomedical Engineering, Granada, Spain, April 20-22, 2016, Springer, 2016, pp. 315--321.

\bibitem{bardini2017using}
R.~Bardini, G.~Politano, A.~Benso, S.~Di~Carlo, {Using multi-level {Petri} nets models to simulate microbiota resistance to antibiotics}, in: BIBM 2017: Int. Conf. on Bioinf. and Biomedicine, 2017, pp. 128--133.
\newblock \href {https://doi.org/10.1109/BIBM.2017.8217637} {\path{doi:10.1109/BIBM.2017.8217637}}.

\bibitem{kummer2000renew}
O.~Kummer, F.~Wienberg, M.~Duvigneau, M.~K{\"o}hler, D.~Moldt, H.~R{\"o}lke, Renew--the reference net workshop, in: Tool Demonstrations, 21st Int. Conf. on Application and Theory of {Petri} Nets, Aarhus, Denmark, 2000, pp. 87--89.

\bibitem{willrodt2020modular}
S.~Willrodt, D.~Moldt, M.~Simon, Modular model checking of reference nets: Momoc., in: PNSE@ {Petri} Nets, 2020, pp. 181--193.

\bibitem{capra2023encoding}
L.~Capra, M.~K{\"o}hler-Bu{\ss}meier, Encoding {Nets-Within-Nets} in maude, in: Science and Information Conference, Springer, 2023, pp. 355--372.
\newblock \href {https://doi.org/http://dx.doi.org/10.1007/978-3-031-37963-5\_25} {\path{doi:http://dx.doi.org/10.1007/978-3-031-37963-5\_25}}.

\bibitem{figat2019methodology}
M.~Figat, C.~Zieli{\'n}ski, {Methodology of designing multi-agent robot control systems utilising hierarchical {Petri} nets}, in: ICRA 2019: Int. Conf. on Robotics and Automation, 2019, pp. 3363--3369.
\newblock \href {https://doi.org/10.1109/ICRA.2019.8794201} {\path{doi:10.1109/ICRA.2019.8794201}}.

\bibitem{baier2008principles}
C.~Baier and J.-P.~Katoen.
\newblock \emph{Principles of Model Checking}.
\newblock MIT Press, 2008.

\bibitem{Clarke99}
E.~M.~Clarke, O.~Grumberg, and D.~Peled.
\newblock \emph{Model Checking}.
\newblock MIT Press, 1999.

\bibitem{esparza2018one}
J.~Esparza, J.~K\v{r}et\'{\i}nsk{\`y}, and S.~Sickert.
\newblock One theorem to rule them all: A unified translation of {LTL} into $\omega$-automata.
\newblock In \emph{Proceedings of the 33rd Annual ACM/IEEE Symposium on Logic in Computer Science (LICS)}, pages 384--393, 2018.

\bibitem{kloetzer2020path}
M.~Kloetzer and C.~Mahulea.  
\newblock Path planning for robotic teams based on {LTL} specifications and {Petri} net models.  
\newblock \emph{Discrete Event Dynamic Systems}, 30(1):55--79, 2020.  
\newblock Springer.

\bibitem{Wolper83}
P.~Wolper, M.~Vardi, A.~Sistla, Reasoning about infinite computation paths, in: E.~N. et~al. (Ed.), Proceedings of the 24th IEEE Symposium on Foundations of Computer Science, Tucson, AZ, 1983, pp. 185--194.
\newblock \href {https://doi.org/10.1109/SFCS.1983.51} {\path{doi:10.1109/SFCS.1983.51}}.

\bibitem{kupferman2001model}
O.~Kupferman, M.~Y. Vardi, Model checking of safety properties, Formal methods in system design 19 (2001) 291--314.
\newblock \href {https://doi.org/http://dx.doi.org/10.1007/3-540-48683-6\_17} {\path{doi:http://dx.doi.org/10.1007/3-540-48683-6\_17}}.

\bibitem{cabasino2011discrete}
M.~P. Cabasino, A.~Giua, M.~Pocci, C.~Seatzu, Discrete event diagnosis using labeled {P}etri nets. an application to manufacturing systems, Control Engineering Practice 19~(9) (2011) 989--1001.
\newblock \href {https://doi.org/https://doi.org/10.1016/j.conengprac.2010.12.010} {\path{doi:https://doi.org/10.1016/j.conengprac.2010.12.010}}.

\bibitem{Jensen1991-ym}
K.~Jensen, Coloured {P}etri nets: A high level language for system design and analysis, in: High-level {P}etri Nets, Springer Berlin Heidelberg, Berlin, Heidelberg, 1991, pp. 44--119.
\newblock \href {https://doi.org/http://dx.doi.org/10.1007/3-540-53863-1\_31} {\path{doi:http://dx.doi.org/10.1007/3-540-53863-1\_31}}.

\bibitem{hollingsworth2015reducing}
K.~G. Hollingsworth, {Reducing acquisition time in clinical MRI by data undersampling and compressed sensing reconstruction}, Physics in Medicine \& Biology 60~(21) (2015) R297.
\newblock \href {https://doi.org/10.1088/0031-9155/60/21/R297} {\path{doi:10.1088/0031-9155/60/21/R297}}.

\bibitem{hennersperger2016towards}
C.~Hennersperger, B.~Fuerst, S.~Virga, O.~Zettinig, B.~Frisch, T.~Neff, N.~Navab, Towards mri-based autonomous robotic us acquisitions: a first feasibility study, IEEE transactions on medical imaging 36~(2) (2016) 538--548.
\newblock \href {https://doi.org/10.1109/TMI.2016.2620723} {\path{doi:10.1109/TMI.2016.2620723}}.

\bibitem{li2021overview}
K.~Li, Y.~Xu, M.~Q.-H. Meng, An overview of systems and techniques for autonomous robotic ultrasound acquisitions, IEEE Trans. on Medical Robotics and Bionics 3~(2) (2021) 510--524.
\newblock \href {https://doi.org/10.1109/TMRB.2021.3072190} {\path{doi:10.1109/TMRB.2021.3072190}}.

\bibitem{Kloetzer2020-ey}
M.~Kloetzer, C.~Mahulea, {Path planning for robotic teams based on {LTL} specifications and {Petri} net models}, Discrete Event Dyn. Syst.: Theory Appl. 30~(1) (2020) 55--79.
\newblock \href {https://doi.org/http://dx.doi.org/10.1007/s10626-019-00300-1} {\path{doi:http://dx.doi.org/10.1007/s10626-019-00300-1}}.

\bibitem{IPGoMaKl15}
R.~Gonz\'alez, C.~Mahulea, M.~Kloetzer, A {Matlab}-based interactive simulator for mobile robotics, in: IEEE CASE'2015: Int. Conf. on Autom. Science and Engineering, 2015.
\newblock \href {https://doi.org/10.1109/CoASE.2015.7294097} {\path{doi:10.1109/CoASE.2015.7294097}}.

\bibitem{cplex2009v12}
I.~I. Cplex, V12. 1: User’s manual for {CPLEX}, International Business Machines Corporation 46~(53) (2009) 157.

\bibitem{GastinOddoux2001}
Paul Gastin and Denis Oddoux.  
\newblock Fast LTL to Büchi Automata Translation.  
\newblock In {\em Proceedings of the 13th International Conference on Computer Aided Verification (CAV '01)}, pages 53--65. Springer-Verlag, Berlin, Heidelberg, 2001.  
ISBN 3540423451.


\end{thebibliography}
\end{document}